# Distributed Lag Transformer based on Time-Variable-Aware Learning for Explainable Multivariate Time Series Forecasting

Younghwi Kim, Dohee Kim, *Member, IEEE*, Joongrock Kim, Sunghyun Sim, *Member, IEEE*

***Abstract*— Time series data is a key element of big data analytics, commonly found in domains such as finance, healthcare, climate forecasting, and transportation. In large-scale real-world settings, such data is often high-dimensional and multivariate, requiring advanced forecasting methods that are both accurate and interpretable. Although Transformer-based models perform well in multivariate time series forecasting (MTSF), their lack of explainability limits their use in critical applications. To overcome this, we propose Distributed Lag Transformer (DLFormer), a novel Transformer architecture for explainable and scalable MTSF. DLFormer integrates a distributed lag embedding and a time-variable-aware learning (TVAL) mechanism to structurally model both local and global temporal dependencies and explicitly capture the influence of past variables on future outcomes. Experiments on ten benchmark and real-world datasets show that DLFormer achieves state-of-the-art predictive accuracy while offering robust, interpretable insights into variable-wise and temporal dynamics. These results highlight DLFormer's ability to bridge the gap between performance and explainability, making it highly suitable for practical big data forecasting tasks.**

***Index Terms*—Distributed Lag Transformer, Distributed Lag Embedding, Explainable Multivariate Time Series Forecasting, Time-Variable-Aware Learning**

## I. INTRODUCTION

TIME series , which consists of sequential observations recorded over time, serves as a fundamental data type in big data analytics across a wide range of domains including finance, healthcare, climate forecasting, and traffic management [1]–[4]. In real-world big data environments**,** time series data is typically multivariate and high-dimensional, encompassing numerous interdependent variables that evolve simultaneously over time [5]–[7]. This complexity gives rise to the critical problem of multivariate time series forecasting (MTSF) in big data systems, which demands the modeling of both short- and long-range temporal dependencies to generate accurate and interpretable predictions at scale [8]–[10].

Traditional approaches to the MTSF problem include statistical models (e.g., Autoregressive Integrated Moving Average (ARIMA) and Vector Autoregression (VAR)) and deep learning-based models (e.g., Recurrent Neural Networks (RNNs), Long Short-Term Memory (LSTM), and Transformers) that have been extensively studied. While statistical models provide explainable structures, they often struggle to capture complex temporal dependencies, especially in high-dimensional data [11]. In contrast, deep learning models have demonstrated remarkable predictive performance but lack explainability, hindering the interpretability of predictions [12]. This limitation is critical in domains where transparency and trust in predictions are essential, such as finance and healthcare [13].

Although deep learning-based approaches such as Transformer models have shown continuous advancement, several critical limitations still.

- ***Lack of explainability for temporal relationships among variable*s**: Transformer-based models leverage attention mechanisms to capture dependencies among variables, yet they do not explicitly model how historical values of individual variables impact future predictions. Consequently, visualizing attention weights is insufficient for quantifying the temporal impact of individual variables, complicating the interpretation of the underlying forecasting dynamics.
- ***Limitations of explainable artificial intelligence (XAI) methods in time series forecasting***: Recent advancements in XAI have primarily targeted non-temporal data domains such as images and text. Conventional XAI techniques, such as Shapley Additive Explanations (SHAP) and Local Interpretable Model-agnostic Explanations (LIME), struggle to incorporate the temporal structure of multivariate time series data, making them inadequate for explaining the lagged effects of variables on future predictions [12].
- ***Trade-off between predictive performance and explainability***: Existing research often prioritized improving predictive accuracy in Transformer-based

This work was partly supported by the Innovative Human Resource Development for Local Intellectualization program through the Institute of Information & Communications Technology Planning & Evaluation (IITP) grant funded by the Korea government (MSIT) (IITP-2024-2020-0-01791, 50%) and by the National Research Foundation of Korea (NRF) grant funded by the Korea government (MSIT) (No.RS-2023-00218913, 50%). (*Co-first author*: Younghwi Kim and Dohee Kim, *Corresponding author*: Sunghyun Sim.) The source code is publicly available at: http://github.com/kYounghwi/DLFormer_official

Younghwi Kim is with the Graduate School of AI Convergence Engineering, Changwon National University, 20 Changwondaehak-ro, Ui-chang Gu, 51140 Changwon, South Korea (e-mail: dudgnl6032@naver.com)

Dohee Kim is with the Safe & Clean Supply Chain Research Center, Pusan National University, 30-Jan-jeon Dong, Geum-Jeong Gu, 46241, Busan, South Korea (e-mail: kimdohee@pusan.ac.kr).

Sunghyun Sim and Jorak Kim are with the School of AI Convergence, Changwon National University, 20 Changwondaehak-ro, Ui-chang Gu, 51140 Changwon, South Korea (e-mail: ssh@changwon.ac.kr, jrkim@changwon.ac.kr).

forecasting models while overlooking rigorous evaluations of explainability. High explainability models commonly demonstrate reduced predictive performance, and effective methods to balance both aspects remain lacking [14].

To address these challenges, we propose a novel approach that enhances the explainability of the MTSF problem while maintaining high predictive performance. Specifically, we introduce time-variable-aware learning (TVAL), which explicitly models the contributions of individual variables' past values (temporal lags) to future predictions.

To enable TVAL, we developed the Distributed Lag Transformer (DLFormer), which integrates the well-established distributed lag mechanism [15] from econometrics into Transformer-based time series models. DLFormer decomposes each variable's temporal information, creating a structured representation of local (within-variable) and global (across-variable) temporal dependencies. The core component, distributed lag embedding (DL Embedding), enables the model to effectively learn and quantify the influence of past observations on future forecasts. Our research offers the following contributions:

- **(C1)** We introduce TVAL, a new learning paradigm for MTSF that explicitly captures variable-specific temporal dependencies, enhancing interpretability and predictive performance.
- **(C2)** We propose DLFormer, the first Transformer-based model that structurally integrates the distributed lag mechanism to model the impact of past observations on future predictions.
- **(C3)** We develop DL Embedding, a novel embedding strategy that encodes local and global temporal relationships in a structured and interpretable representation.
- **(C4)** Extensive experiments on ten benchmark and real-world datasets demonstrate that DLFormer achieves state-of-the-art predictive accuracy while significantly enhancing model explainability from temporal and variable perspectives.

The remainder of this paper is organized as follows. Section II reviews related literature. Section III details the proposed DLFormer model. Section IV presents comparative experiments against various state-of-the-art methods, while Section V validates the effectiveness of DLFormer through extensive ablation studies. Finally, Section VI concludes the paper by summarizing the findings and exploring future research directions.

## II. Related Work

This section reviews XAI methods and discusses research that apply these approaches to multivariate time-series data.

### *A. Explainable Artificial Intelligence*

With the increasing adoption of machine learning and deep learning models across various industries, the demand for XAI has grown significantly [16]. The "black-box problem" in deep learning [17], where models produce highly accurate results but lack transparency in decision-making, has spurred extensive research in XAI [18]. The core objective of XAI is to provide understandable explanations for AI decisions, enabling stakeholders to trust and validate model outputs [12]. XAI methods can be categorized across three independent dimensions: the scope of explanation, timing of explanation, and methodological dependency [11], [12].

First, regarding the scope of explanation, XAI approaches can be divided into global and local Explanations. Global explanations aim to understand the model's overall decision-making process, deriving general rules and patterns applicable to the entire dataset. Analyzing feature importance and decision pathways allows researchers and practitioners to assess whether the model behaves logically and consistently [19]. In contrast, local explanations aim to explain individual predictions by analyzing how specific inputs led to specific outputs. This is crucial in high-stakes domains such as medical diagnostics and fraud detection, where understanding the rationale behind individual decisions is critical [18], [20]. Modern XAI research seeks to balance both explanations, but the complexity of deep learning models presents challenges, resulting in many methods focusing on one type over the other [21].

Second, based on the timing of explanation, XAI methods are categorized into post-hoc and ante-hoc approaches. Post-hoc methods provide explanations after model training, making them independent of the underlying architecture. These methods are widely used because they introduce interpretability to existing models without altering their internal structure [12]. In contrast, ante-hoc methods are designed with transparency as a core principle. Often referred to as White-box models, these methods ensure that decision-making processes are inherently interpretable without additional tools [22].

Third, from the perspective of methodological dependency, XAI methods can be classified into model-agnostic and model-specific techniques. Model-agnostic methods, such as SHAP [23], LIME [24], and Gradient-based approaches, can be applied universally across different machine learning models. In contrast, model-specific methods are tailored for specific model architectures, particularly deep learning models such as Gradient-weighted Class Activation Mapping (Grad-CAM) [25] and Grad-CAM++ [26]. These methods leverage internal model information to generate more precise and architecture-aware explanations but lack generalizability across various datasets and tasks [27].

### *B. Explainable Method for Time-series forecasting*

Time-series forecasting involves nonlinear temporal dependencies and delayed interactions among multiple variables, complicating the development and evaluation of explainable methods. Existing XAI methods, such as LIME and SHAP, were originally designed for models handling cross-sectional data, limiting their direct application to time-series tasks [12], [28]. Furthermore, forecasting requires more intricate dynamics to be explained compared to classification problems, as it models long-term dependencies and interactions over time [29]. Consequently, recent research has focused on adapting existing XAI methods to better suit time-series forecasting or specifically designing new ante-hoc approaches [28]-[31].

TABLE I
EXPLAINABLE TIME-SERIES FORECASTING MODELS

| Model | Methodology | Ante-hoc/ Post-hoc | Input Series (Multi / Uni) | Explainable Dimension (Multi / Uni) | Evaluation (Global / Local) | Evaluation (Qualitative / Quantitative) |
|---|---|---|---|---|---|---|
| IMV-LSTM (2017) | Attention | Ante-hoc | M | M | Global | Qual / Quant |
| TFT (2019) | Attention / VSN | Ante-hoc | M | M | Global | Qual |
| TACN (2020) | Attention | Ante-hoc | M | U | Local | Qual |
| XCM (2021) | Grad-Cam | Poct-hoc | M | M | Local | Qual / Quant |
| NLinear (2023) | Linear | Ante-hoc | U | U | Global | Qual |
| MTEX (2024) | Integrated-Gradient | Poct-hoc | M | M | Local | Qual / Quant |
| iTransformer (2024) | Attention | Ante-hoc | M | U | Global | Qual |
| **Ours (DLFormer)** | **Attention** | **Ante-hoc** | **M** | **M** | **Global / Local** | **Quali / Quant** |

Explainable time-series forecasting methods can be classified based on their emphasis on variable dimension (variable importance (VI)) or temporal dimension (temporal importance (TI)) [32]. Accordingly, explainability methods can be divided into uni-dimensional explainable (UDE) models, providing explanations for one dimension (feature or time), and multi-dimensional explainable (MDE) models, which considers both dimensions simultaneously. UDE methods primarily analyze specific components of a model, often emphasizing predictive performance over explainability. In contrast, MDE methods aim to provide a more comprehensive explanation by modeling relationships between time and variables; however, they entail increased computational complexity, which can impact predictive performance.

Notable UDE methods include Temporal Attention Convolutional Neural Networks (TACN) [30], iTransformer [33], and NLinear [34]. TACN applies attention mechanisms to time tokens to highlight TI, integrating this with Temporal Convolutional Networks to form a structured model. iTransformer reverses the standard transformer input structure, treating time and variable dimensions interchangeably, thereby capturing variable interactions using attention mechanisms. NLinear employs explicit autoregressive linear layers for each time series, effectively modeling autoregressive temporal patterns. Although these methods enhance predictive accuracy, they often fail to provide explanations that address time and variable dependencies simultaneously.

In contrast, MDE methods integrate feature and temporal dependencies into a unified interpretability framework. Representative models include Interpretable Multi-Variable (IMV)-LSTM [35] and Temporal Fusion Transformer (TFT) [29]. IMV-LSTM extends traditional LSTMs by incorporating a mixture attention mechanism, enabling dynamic capture of both mechanisms during prediction. In contrast, TFT employs attention-based TI estimation while integrating a variable selection network (VSN) within the model, providing separate evaluations for VI and TI. Additionally, post-hoc explainability methods have been adapted for time-series forecasting. For instance, Grad-CAM has been applied to custom designed CNN-based architectures for time-series tasks (eXplainable CNN method for MTS, XCM) [31], while more recently, Integrated Gradients has been demonstrated to be effective for CNN-based structural networks (Multivariate Time series Explanations for Predictions, MTEX) [28] in time-series forecasting. Although MDE methods offer more robust and comprehensive explanations than UDE methods, they must balance the challenge of modeling complex temporal dynamics while maintaining interpretability. Therefore, their predictive performance is often lower than that of UDE methods, which are optimized for forecasting accuracy. Table I summarizes the characteristics of the previously defined explainable time-series forecasting models alongside our model.

## III. DISTRIBUTED LAG TRANSFORMER

This session introduces the novel explainable MTSF model DLFormer and provides a detailed explanation of the TVAL method within it. Table II summarizes the notations used, while Fig. 1 illustrates the overall architecture of the proposed DLFormer. DLFormer comprises three main components: (1) DL Embedding, (2) Distributed Lag Encoder (DL Encoder), and (3) Distributed Lag Decoder (DL Decoder).

### A. Distributed Lag Embedding

The DL Embedding module embeds input data to capture inter-variable dependencies within a multivariate time series and temporal dependencies in individual time series, representing them in the embedding space. This is achieved by sequentially concatenating the $F$ univariate time series components from the multivariate dataset, resulting in the following input data formulation:

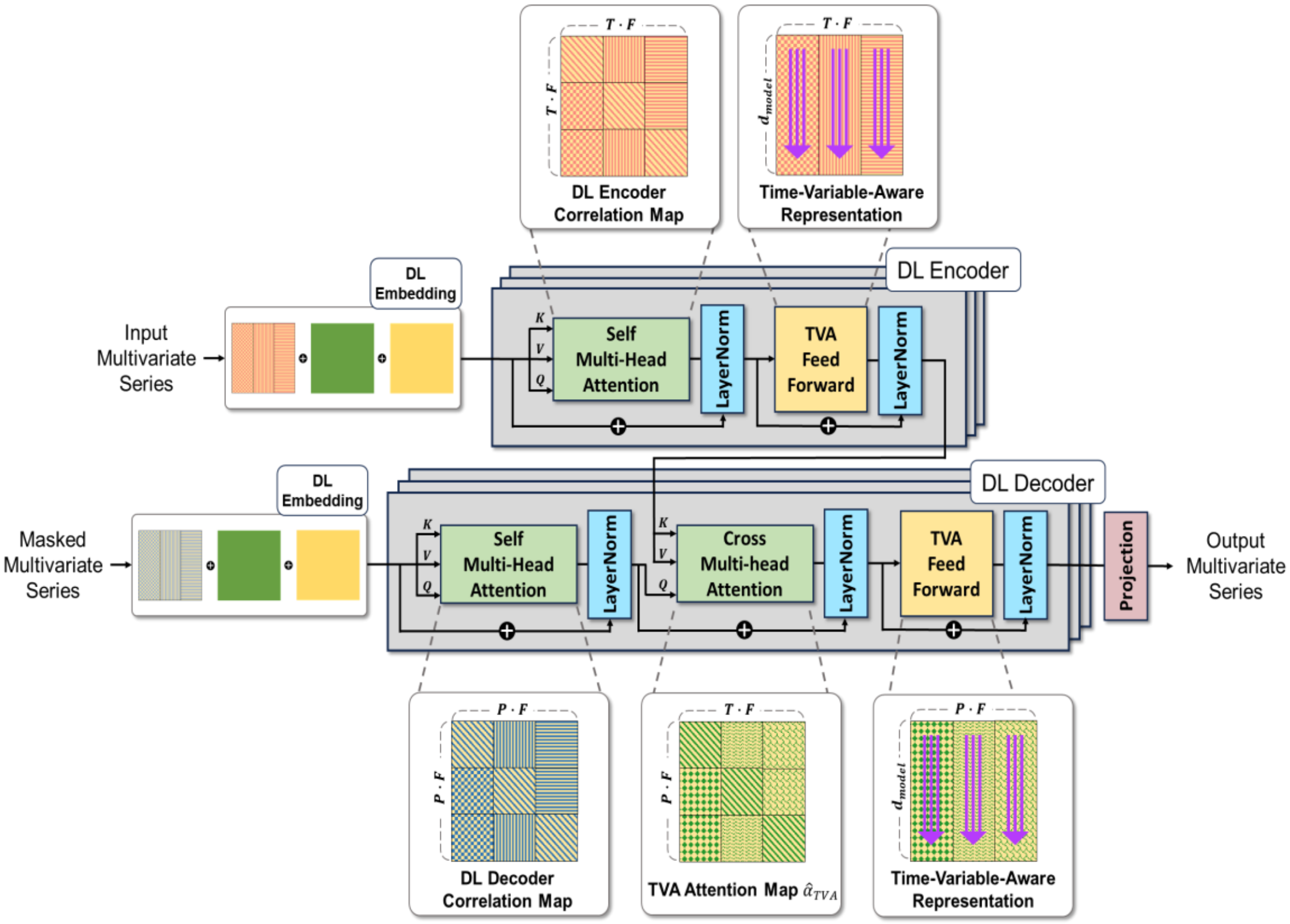

**Fig. 1.** Proposed DLFormer network structure.

TABLE II
SUMMARY OF NOTATIONS

| Symbols | Description |
|---|---|
| $S, F$ | Number of time series samples and variables, respectively |
| $T, P$ | Input sequence length and forecasting horizon |
| $N, M, d$ | Number of encoder layer, decoder layer, and embedding dimension |
| $X_{1:T} \in \mathbb{R}^{S \times T \times F}$ | Input multivariate time series with $T$ sequence |
| $X_{T:T+P} \in \mathbb{R}^{S \times P \times F}$ | Target sequence representing the $P$ future observations |
| $X_{DL} \in \mathbb{R}^{S \times T \cdot F \times 1}$ | Input to the distributed lag embedding |
| $H_{DL} \in \mathbb{R}^{S \times T \cdot F \times d_{model}}$ | Embedded representation of $X_{DL}$ |
| $z^{enc} \in \mathbb{R}^{S \times T \cdot F \times d_{model}}$ | Encoder output |
| $z^{dec} \in \mathbb{R}^{S \times T \cdot F \times d_{model}}$ | Decoder output |
| $\hat{\alpha}$ | Explanation generated by DLFormer for time-series |
| $\ell, \varphi, \nabla_{\varphi}$ | Loss function, model parameters, and gradient of $\varphi$, respectively |

$$X_{DL} = \{x_1^1, \cdots, x_T^1, \cdots, x_1^F, \cdots, x_T^F\}. \quad (1)$$

Here, a linear transformation is performed from $X_{DL} \in \mathbb{R}^{S \times T \cdot F \times 1} \rightarrow H_{DL} \in \mathbb{R}^{S \times T \cdot F \times d_{model}}$, where $d_{model}$ represents the embedding dimension. Subsequently, to inject local positional information into each univariate time series and global positional information across the entire sequence, we introduce local sequence position embedding (LSPE) and global sequence position embedding (GSPE), both with dimensionality matching that of $H_{DL}$. LSPE and GSPE are derived from the sequence position embedding (SPE), defined as follows:

$$SPE_{t,d} = \begin{cases} \sin\left(t / C^{d/d_{model}}\right), if\ d\ mod\ 2 = 0 \\ \cos\left(t / C^{d/d_{model}}\right), \quad otherwise \end{cases}. \quad (2)$$

Here, $t \in \{1, \dots, T\}$represents the position within the input sequence, while $d \in \{0, \dots, d_{model}\}$ denotes the index of the embedding dimension. $C$ is a constant that determines the periodicity. By alternately applying sine and cosine functions with different frequencies to the embedding, a unique positional embedding value is added to each sequence.

LSPE is structured as a combination of $F$ SPEs, each encoding sequential information of an individual univariate time series, allowing local sequential information injection into the embedding vector. The LSPE is defined as follows:

$$LSPE_{t,d} = Concat(SPE_{t,d}^1, \cdots, SPE_{t,d}^F). \quad (3)$$

In contrast, GSPE constructs an SPE with the same dimensionality as the linearly transformed representation $H_{DL}$, by defining the range of $t \in \{1, \dots, T \cdot F\}$. Consequently, the final DL Embedding is defined as follows:

$$DLE = H_{DL} + GSPE_{p,d} + LSPE_{p,d}. \quad (4)$$

Fig. 2 illustrates the process of applying DL Embedding to a multivariate time series dataset consisting of three univariate time series. First, the multivariate time series data, composed of three univariate time series, is rearranged into a single-dimensional sequence to construct $X_{DL}$. Next, a linear transformation with dimensionality $d_{model}$ is applied to $X_{DL}$ obtaining $H_{DL}$ which has a dimensionality of $d_{model} \times T \cdot 3$. LSPE is obtained by combining three SPEs, each with a dimensionality of $d_{model} \times T$. In contrast, GSPE is constructed as a single SPE with dimensionality $d_{model} \times T \cdot 3$. Finally, the embedded vector is obtained by summing $H_{DL}$, LSPE, and GSPE.

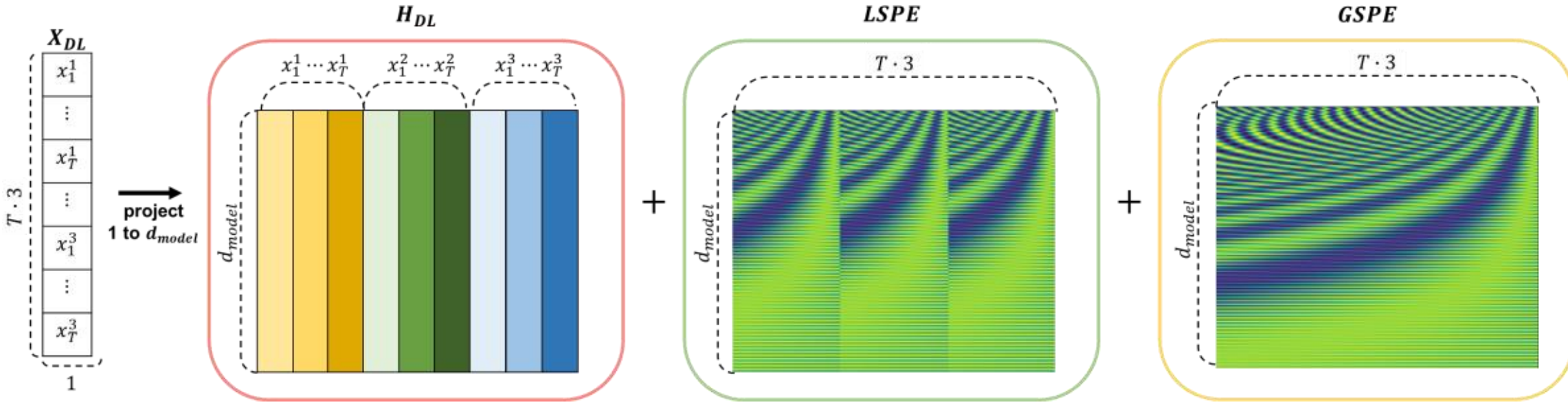

**Fig. 2.** Detailed visualization of Distributed Lag Embedding operation.

### *B. Distributed Lag Encoder and Distributed Lag Decoder*

The DL Encoder and DL Decoder combine a standard Transformer with DL Embedding. The model consists of $N$ encoders and $M$ decoders, each learning through the following attention mechanism:

$$A(Q,K) = Softmax\left(QK^{\mathrm{T}}/\sqrt{d_{attn}}\right), \quad (5)$$

$$Attention(Q,K,V) = A(Q,K)V, \quad (6)$$

$$H_i = Attention(QW_i^Q, KW_i^K, VW_i^V). \quad (7)$$

In (7), inputs $Q$, $K$, and $V$ denote the DL Embedding vectors, each with an embedding dimension of $d_{model}$. The parameter matrices $W_i^Q, W_i^K, W_i^V$ correspond to the $i^{th}$ attention head, applied to $Q$, $K$, and $V$, respectively. The output $H_i$ represents the result from the $i^{th}$ attention head. To learn from multiple perspectives, multi-head attention (MHA) performs the operation $I$ times, following linear transformation of the input embedding dimension to $d_{attn}$, as follows:

$$MHA(Q,K,V) = Concat(H_1, \cdots, H_I)W^M. \quad (8)$$

The DL Encoder utilizes the DL Embedding vector as the input to Self-MHA for learning. This process is formulated as follows:

$$z_{n,1}^{enc} = MHA\left(z_{n-1,2}^{enc}, z_{n-1,2}^{enc}, z_{n-1,2}^{enc}\right) + z_{n-1,2}^{enc}, \quad (9)$$

$$z_{n,2}^{enc} = TVAFF\left(z_{n,1}^{enc}\right) + z_{n,1}^{enc}. \quad (10)$$

Here, $z_{n,2}^{enc}$ ($1 \leq n \leq N$) represents the output of the nth encoder, which serves as the input for the n+1th encoder. For the first encoder, the input $z_{0,2}^{enc}$ is set to $DLE$, as in (4).

The time-variable-aware (TVA) feed forward consists of two linear transformations with an activation function in between. TVA feed forward enables the learning of intrinsic characteristics by embedding the representations of each variable at each time step through dense nonlinear connections in multivariate time series. The DL Decoder is structured as follows:

$$z_{m,1}^{dec} = MHA\left(z_{m-1,3}^{dec}, z_{m-1,3}^{dec}, z_{m-1,3}^{dec}\right) + z_{m-1,3}^{dec}, \quad (11)$$

$$z_{m,2}^{dec} = MHA\left(z_{m,1}^{dec}, z_{N,2}^{enc}, z_{N,2}^{enc}\right) + z_{m,1}^{dec}, \quad (12)$$

$$z_{m,3}^{dec} = TVAFF\left(z_{m,2}^{dec}\right) + z_{m,2}^{dec}. \quad (13)$$

Here, $z_{m,3}^{dec}(1 \leq m \leq M)$ represents the output of the $m^{th}$ decoder, which serves as the input for the subsequent decoder. For the DL Embedding input of the initial decoder $z_{0,3}^{dec}$, we use a zero-masked vector of future time steps $\boldsymbol{X}_{T:T+P}^{0}$. As formulated in (11), the decoder's attention mechanism applies Self-MHA to $z_{m-1,3}^{dec}$; subsequently, it applies Cross-MHA as in (12), where the output from the previous Self-MHA serves as Q, while the final encoder output serves as K and V. The

---

**Algorithm 1** Training procedure

---

    **Input:** Time series data $X_{1,T}, X_{T:T+P}^{0}$

1: $z_{0,2}^{enc} \leftarrow DLE$ applied to $X_{1,T}$

2: $z_{0,3}^{dec} \leftarrow DLE$ applied to $X_{T:T+P}^{0}$

3: **for** *epoch* $= 1, \ldots, MaxEpoch$ **do**

4:     **for** $n = 1, \ldots, N$ **do**     ⊳ DL Encoder

5:         $z_{n,1}^{enc} \leftarrow MHA\left(z_{n-1,2}^{enc}, z_{n-1,2}^{enc}, z_{n-1,2}^{enc}\right) + z_{n-1,2}^{enc}$

6:         $z_{n,2}^{enc} \leftarrow TVAFF\left(z_{n,1}^{enc}\right) + z_{n,1}^{enc}$

7:     **end for**

8:     **for** $m = 1, \ldots, M$ **do**     ⊳ DL Decoder

9:         $z_{m,1}^{dec} \leftarrow MHA\left(z_{m-1,3}^{dec}, z_{m-1,3}^{dec}, z_{m-1,3}^{dec}\right) + z_{m-1,3}^{dec}$

10:         $z_{m,2}^{dec} \leftarrow MHA\left(z_{m,1}^{dec}, z_{N,2}^{enc}, z_{N,2}^{enc}\right) + z_{m,1}^{dec}$

11:         $z_{m,3}^{dec} \leftarrow TVAFF\left(z_{m,2}^{dec}\right) + z_{m,2}^{dec}$

12:     **end for**

13:     $\hat{X}_{T:T+P} = Projection\left(z_{M,3}^{dec}\right)$

14:     $Loss_{\varphi} \leftarrow \ell\left(X_{T:T+P}, \hat{X}_{T:T+P}\right)$

15:     $\varphi \leftarrow \varphi - \eta\nabla_{\varphi}Loss_{\varphi}$     ⊳ Weight update

16: **end for**

---

decoder generates the final output by embedding the variable-wise temporal representations through TVA feed forward as in (13).

The above steps define the encoding and decoding process of variable-wise temporal representation, where DL embedding, structured attention, and TVA feed forward are combined within the DLFormer, forming the TVAL method. This approach offers higher expressive power than conventional Transformer-based models that embed variables or time tokens. Finally, future predictions are derived through a single linear transformation, as follows:

$$\hat{X}_{T:T+P} = Projection\left(z_{M,3}^{dec}\right). \tag{14}$$

Algorithm 1 presents the pseudocode for training the DLFormer.

*C. Time-Variable-Aware Attention Map*

In the DLFormer decoder, Cross-MHA is applied as formulated in (12), where $Q = z_{m,1}^{dec}, K = z_{N,2}^{enc}, V = z_{N,2}^{enc}$. This mechanism enables learning relationships between the predicted values and DL Embedding representations from the encoder. Consequently, the learned relationship, represented by $A\left(z_{m,1}^{dec}, z_{N,2}^{enc}\right) \in \mathbb{R}^{P\cdot F \times T\cdot F}$ in (5), can be interpreted as a saliency attention map highlighting the importance of individual variable time series within the forecasting horizon. We refer to this as the TVA attention map $\hat{\alpha}$. Utilizing the final $M^{\text{th}}$ decoder here, we can formalize it as follows:

$$\hat{\alpha}_{TVA} = \frac{1}{I}\sum_i A_i(z_{M,1}^{dec}, z_{N,2}^{enc}), \tag{15}$$

$$\hat{\alpha}_{Time} = reshape\left(\frac{1}{PF}\sum_{pf}\hat{\alpha}_{TVA}\right), \tag{16}$$

$$\hat{\alpha}_{Var} = \frac{1}{T}\sum_t \hat{\alpha}_{Time}. \tag{17}$$

Here, $A_i$ represents the result from the $i^{\text{th}}$ head's decoder as defined in (5). Therefore, $\hat{\alpha}_{TVA} \in \mathbb{R}^{P\cdot F \times T\cdot F}$ represents the detailed TI of input variables across each forecasting horizon. To simplify, we further process this from the perspectives of TI per variable and VI. After averaging $\hat{\alpha}_{TVA}$ over the forecasting horizon $P \cdot F$, it is reshaped from $\mathbb{R}^{1\times T\cdot F} \to \mathbb{R}^{F\times T}$ resulting in $\hat{\alpha}_{Time} \in \mathbb{R}^{F\times T}$ (16). This is then averaged across the input

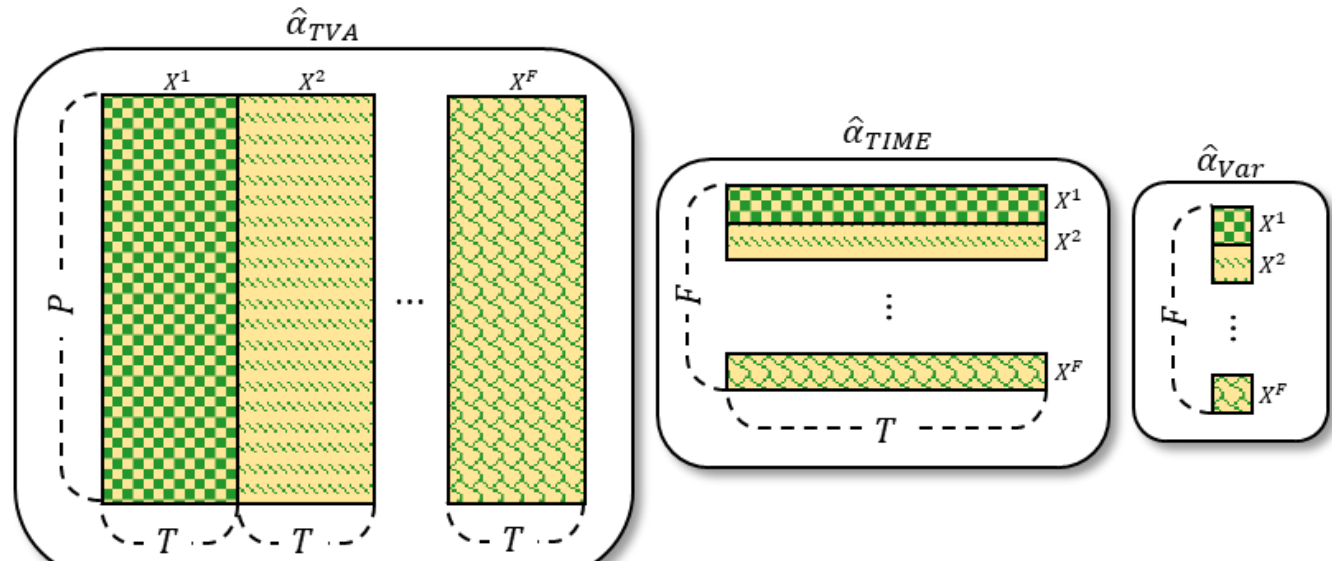

**Fig. 3.** Visualization example of the time-variable-aware attention map.

length dimension $T$ to obtain $\hat{\alpha}_{Var} \in \mathbb{R}^F$ (17).

A visualization for a predicted single variable is provided in Fig. 3. $\hat{\alpha}_{TVA}$ represents the TI distribution of each input variable arranged according to the forecast horizon. The value of $\hat{\alpha}_{Time}$ is averaged based on the forecast horizon and $\hat{\alpha}_{Var}$ is further averaged over the input sequence to provide explanations for predictions at each level. Thus, $\hat{\alpha}_{TVA}$, $\hat{\alpha}_{Time}$, and $\hat{\alpha}_{Var}$ can be expressed from three different perspectives, depending on the required level of explanation for each dataset's domain. This is possible because the TVA attention map captures the importance of time and variables in multivariate time series inputs, facilitating the extraction of accurate insights from the MTSF problem. Algorithm 2 provides the pseudocode of the inference and explanation extraction process for DLFormer.

**Algorithm 2.** Inference and extract explanation procedure

**Input:** Time series data $X_{1,T}, X_{T:T+P}^{0}$
**Output:** Predicted timeseries $\hat{X}_{T:T+P}$, explanation $\hat{\alpha}_{TVA}$, $\hat{\alpha}_{Time}$, and $\hat{\alpha}_{Var}$
1: $z_{0,2}^{enc} \leftarrow DLE$ applied to $X_{1,T}$
2: $z_{0,3}^{dec} \leftarrow DLE$ applied to $X_{T:T+P}^{0}$
3: **for** $n = 1,\ldots, N$ **do**
4: $z_{n,1}^{enc} \leftarrow MHA\left(z_{n-1,2}^{enc}, z_{n-1,2}^{enc}, z_{n-1,2}^{enc}\right) + z_{n-1,2}^{enc}$
5: $z_{n,2}^{enc} \leftarrow TVAFF\left(z_{n,1}^{enc}\right) + z_{n,1}^{enc}$
6: **end for**
7: **for** $m = 1,\ldots, M$ **do**
8: $z_{m,1}^{dec} \leftarrow MHA\left(z_{m-1,3}^{dec}, z_{m-1,3}^{dec}, z_{m-1,3}^{dec}\right) + z_{m-1,3}^{dec}$
9: $z_{m,2}^{dec} \leftarrow MHA\left(z_{m,1}^{dec}, z_{N,2}^{enc}, z_{N,2}^{enc}\right) + z_{m,1}^{dec}$
10: $z_{m,3}^{dec} \leftarrow TVAFF\left(z_{m,2}^{dec}\right) + z_{m,2}^{dec}$
11: **end for**
12: $\hat{X}_{T:T+P} = Projection\left(z_{M,3}^{dec}\right)$
13: $\hat{\alpha}_{TVA} = \frac{1}{I}\sum_i A_i(z_{M,1}^{dec}, z_{N,2}^{enc})$
14: $\hat{\alpha}_{Time} = reshape\left(\frac{1}{PF}\sum_{pf}\hat{\alpha}_{TVA}(p)\right)$
15: $\hat{\alpha}_{Var} = \frac{1}{T}\sum_t \hat{\alpha}_{Time}(t)$

## IV. EXPERIMENTS

*A. Dataset and Experimental Settings*

We conducted comprehensive experiments to evaluate and analyze the predictive performance and explainability of DLFormer. For comparative analysis, we utilized one proprietary multivariate time-series dataset (Port Air Quality (PAQ)) and nine publicly available datasets (Volatility, PM2.5, SML, Exchange, Weather, ETTh1, ETTh2, ETTm1, and ETTm2). Table IV provides detailed descriptions and variable information for each dataset.

All datasets were divided into training, validation, and test sets with a 70%, 10%, and 20% split, respectively. Experiments were conducted on long-term (96, 192, 336, and 720) and short-term (3, 6, 12, and 36) forecasting tasks to evaluate the model's performance. Input sequence lengths were set to 48 and 36 for long- and short-term forecasting.

TABLE IV
DETAILS OF TIME-SERIES DATASETS USED IN THE EXPERIMENTS

| Dataset | Number of Variables | Data Description | Number of Rows | Collected Period | Frequency |
|---|---|---|---|---|---|
| PAQ | 7 | Air Quality Datasets Collected from Busan Port (Non-Public) | 9357 | 2004.03.01–2005.04.01 | Hourly |
| SML | 16 | Public Datasets Collected from Home Monitoring Systems, used in [35], [37] | 4137 | 2012.03.01–2014.05.01 | 15-minute |
| Volatility | 8 | Public dataset containing daily realized volatility values of the S&P 500 stock index, calculated from intraday data, along with its daily returns, used in [29]. | 4641 | 2000.01.01–2018.06.01 | Daily |
| PM2.5 | 7 | Public Meteorological Datasets Related to Fine Dust Collected in Beijing, China, used in [35] | 33405 | 2010.01.01–2014.01.01 | Hourly |
| Exchange | 8 | Public Exchange Rate Datasets for Specific Countries, used in the Darts Library [38] | 7588 | 1990.01.01–2016.01.01 | Daily |
| Weather | 13 | Public Outdoor Weather-Related Datasets Collected in Germany, used in the Darts Library [38] | 52704 | 2020.01.01–2020.12.01 | 10-minute |
| ETTh1, 2 | 7 | Public Datasets for Power Transformers in Substations, used in the Darts Library [38] | 17420 | 2016.07.01–2018.06.01 | Hourly |
| ETTm1, 2 | 7 | Public Datasets for Power Transformers in Substations, used in the Darts Library [38] | 69680 | 2016.07.01–2018.06.01 | 15-minute |

**Baseline** We selected seven explainable forecasting models, including UDE and MDE methods. The UDE methods used were TACN, NLinear, and iTransformer; MDE methods include IMV-LSTM, XCM, MTEX, and TFT.

**Implementation Details** The seven benchmark models were evaluated using their official GitHub implementations. Table III presents the parameter setting ranges for each model.

The optimal parameters that yielded the best performance were selected and used in the comparative experiment. The L2 loss function was employed during training with a fixed batch size of 32. The ADAM [36] optimizer was used for weight updates over 1000 training epochs. The model demonstrating the best performance on the validation dataset was applied to the test dataset, and the results were recorded. Additionally, to evaluate the average forecasting performance of each model, we conducted 10 individual training sessions were conducted. Although MTEX and XCM were originally developed for multivariate time-series classification, this study adapted both for forecasting to enable a fair comparison with post-hoc explainability methods under MDE methods. Specifically, we replaced the class count in the projection layer with the forecasting horizon and removed the Softmax activation function.

*B. Evaluation Metrics*

We compared our model with each baseline model based on 1) forecasting performance and 2) explainability performance. For predictive performance, we employed three evaluation metrics: mean squared error (MSE), mean absolute error (MAE), and correlation coefficient (COR). Their mathematical definitions are as follows:

$$MSE = \frac{1}{S}\sum_{s=1}(X - \hat{X})^2, \tag{15}$$

$$MAE = \frac{1}{S}\sum_{s=1}\left|X - \hat{X}\right|, \tag{16}$$

$$COR = \frac{\sum_{s=1}(X-\bar{X})(\hat{X}-\bar{\hat{X}})}{\sqrt{\sum_{s=1}(X-\bar{X})^2}\sqrt{\sum_{s=1}^{S}(\hat{X}-\bar{\hat{X}})^2}}. \tag{17}$$

Here, $s$ denotes the index of time series sample $S$. Lower MSE and MAE values indicate more accurate predictions, while a higher COR value signifies better predictive performance. For explainability performance, we assessed the models from two perspectives: quantitative evaluation (robustness) and qualitative evaluation (validity). Quantitative evaluation involved the following steps: 1) The trained forecasting model extracts post-hoc feature importance scores from a variable-level perspective. 2) Variability of the extracted importance scores were assessed using percentage-based metrics (e.g., standard deviation and coefficient of variation) and rank-based metrics (e.g., Kendall's tau and COR). Inconsistency in VI scores under identical experimental conditions indicates stability in the model's extraction process. For qualitative evaluation, the following procedure was applied: 1) The trained model extracts post-hoc feature importance scores considering temporal and variable perspectives. 2) These scores are compared against domain knowledge and prior studies associated with each dataset to assess the validity and interpretability of the model's explanations.

TABLE III
DETAILS OF TIME-SERIES DATASETS USED IN THE EXPERIMENTS

| Models | Hyperparameter | Range |
|---|---|---|
| DLFormer | Number of Encoder, Decoder | {1, 2, 3} |
| | Size of embedding dimension | {128, 256, 512} |
| | Number of attention head | {2, 4, 8} |
| iTransformer | Number of Encoder | {1, 2, 3} |
| | Size of hidden dimension | {128, 256, 512} |
| NLinear | No Hyperparameter | - |
| TFT | Number of LSTM layer | {2, 3, 4} |
| | Size of hidden dimension | {64, 128, 256} |
| | Number of attention head | {4, 6, 8} |
| TACN | Number of TCN layer | {5, 6, 7} |
| | Size of TCN filter | {32, 64, 128} |
| IMV-LSTM | Size of hidden dimension | {64, 128, 256} |
| MTEX | Size of CNN filter | {64. 128, 256} |
| XCM | Size of CNN filter | {64. 128, 256} |

### *C. Experimental Results*

**Forecasting Performance** Tables V and VI present the overall results of the long-term forecasting experiments, while Tables VII and VIII show the results for the short-term forecasting tasks. All values are averaged over 10 independent runs. The best performance for each dataset and forecasting length is highlighted in bold red, whereas the second-best performance is blue underline. Due to dataset size limitations, forecasting experiments of length 720 were not conducted for the SML and Volatility datasets.

MDE models exhibit lower forecasting performance compared to UDE models. In contrast, DLFormer consistently outperforms or matches UDE methods. In Table V, DLFormer achieves the best results most frequently, with the previously recognized state-of-the-art model, iTransformer, ranking second, followed by NLinear. In Table VI, DLFormer shows relatively weaker performance on the Weather, ETTh1, and ETTm1 datasets but delivers the best performance on the ETTh2 and ETTm2 datasets.

In long-term forecasting, DLFormer demonstrates significant improvements in average MSE over other models: 4.03 over iTransformer, 10.58 over NLinear, 49.12 over TFT, 152.14 over TACN, 500.25 over IMV-LSTM, 377.58 over MTEX, and 287.15% over XCM.

Similarly, in short-term forecasting (Table VII), DLFormer and iTransformer show comparable performance, consistently outperforming the other models. DLFormer achieves the highest number of best results, followed closely by NLinear. In Table VIII, DLFormer displays relatively weaker performance on the ETTh1 and ETTm1 datasets but outperforms other models on the remaining datasets, showing comparable performance to iTransformer. DLFormer improves average MSE by 0.79 over iTransformer, 17.09 over NLinear, 29.06 over TFT, 20.34 over TACN, 1143.59 over IMV-LSTM, 968.38 over MTEX, and 89.74 over XCM in short-term forecasting tasks.

Table IX presents the weighted average results for each dataset, ranked by long-term and short-term forecasting horizons, to quantitatively identify the best-performing models. The results confirm that DLFormer achieves the highest performance in long-term forecasting. The iTransformer ranks second, followed by NLinear, XCM, TFT, TACN, MTEX, and IMV-LSTM, in that order. In short-term forecasting, iTransformer ranks highest, closely followed by DLFormer, consistently outperforming the other models. Subsequent ranks include NLinear, TFT, TACN, IMV-LSTM, XCM, and MTEX, with the remaining models showing comparable performance

TABLE V

RESULTS OF COMPARATIVE EXPERIMENTS ON LONG-TERM TIME SERIES FORECASTING

| Dataset | | PAQ | | | SML | | | Volatility | | | PM2.5 | | | Exchange | | |
|---|---|---|---|---|---|---|---|---|---|---|---|---|---|---|---|---|
| Metrics | | MSE | MAE | COR | MSE | MAE | COR | MSE | MAE | COR | MSE | MAE | COR | MSE | MAE | COR |
| Ours | 96 | **1.032** | **0.708** | **0.673** | 0.292 | 0.407 | 0.903 | 0.303 | 0.412 | 0.309 | **0.669** | **0.607** | **0.264** | **0.058** | **0.186** | **0.267** |
| | 192 | **1.072** | **0.735** | **0.633** | 0.445 | 0.513 | 0.795 | **0.549** | **0.574** | **0.360** | **0.716** | **0.630** | **0.168** | 0.147 | 0.285 | 0.290 |
| | 336 | **1.162** | **0.757** | **0.616** | **0.477** | **0.550** | 0.755 | 1.552 | 0.900 | 0.374 | **0.737** | **0.643** | **0.123** | 0.266 | 0.401 | 0.276 |
| | 720 | **1.136** | **0.759** | 0.525 | - | - | - | - | - | - | **0.750** | **0.651** | **0.069** | **0.256** | **0.399** | **0.688** |
| iTransformer | 96 | 1.148 | 0.750 | 0.594 | 0.255 | 0.403 | **0.920** | **0.285** | 0.409 | 0.157 | 0.878 | 0.673 | 0.207 | 0.069 | 0.199 | 0.029 |
| | 192 | 1.191 | 0.783 | 0.555 | 0.381 | 0.509 | **0.859** | 0.644 | 0.630 | 0.125 | 0.936 | 0.701 | 0.144 | **0.135** | **0.277** | 0.000 |
| | 336 | 1.203 | 0.796 | 0.548 | 0.566 | 0.611 | **0.798** | 1.426 | 0.902 | 0.084 | 0.961 | 0.705 | 0.099 | 0.245 | 0.375 | -0.030 |
| | 720 | 1.176 | 0.815 | **0.546** | - | - | - | - | - | - | 0.856 | 0.672 | 0.064 | 0.725 | 0.677 | -0.020 |
| NLinear | 96 | 1.281 | 0.827 | 0.545 | **0.249** | **0.399** | 0.916 | 0.291 | **0.406** | 0.211 | 0.947 | 0.700 | 0.181 | 0.075 | 0.208 | -0.070 |
| | 192 | 1.319 | 0.846 | 0.515 | **0.372** | **0.497** | 0.857 | 0.628 | 0.615 | 0.000 | 1.030 | 0.742 | 0.107 | 0.151 | 0.293 | -0.110 |
| | 336 | 1.285 | 0.834 | 0.522 | 0.568 | 0.604 | 0.786 | **1.399** | **0.890** | -0.020 | 1.028 | 0.742 | 0.072 | 0.282 | 0.402 | -0.140 |
| | 720 | 1.289 | 0.857 | 0.497 | - | - | - | - | - | - | 0.884 | 0.697 | 0.056 | 0.788 | 0.692 | -0.080 |
| TFT | 96 | 1.349 | 0.871 | 0.585 | 0.814 | 0.748 | 0.017 | 0.298 | 0.418 | 0.025 | 0.756 | 0.630 | 0.139 | 0.088 | 0.231 | 0.058 |
| | 192 | 1.405 | 0.915 | 0.486 | 0.991 | 0.817 | 0.000 | 0.612 | 0.613 | 0.234 | 0.784 | 0.641 | 0.078 | 0.173 | 0.322 | -0.060 |
| | 336 | 1.344 | 0.883 | 0.550 | 1.131 | 0.862 | -0.030 | 3.064 | 1.293 | -0.470 | 0.797 | 0.644 | 0.051 | **0.203** | **0.349** | 0.226 |
| | 720 | 1.476 | 0.881 | 0.443 | - | - | - | - | - | - | 0.766 | 0.658 | 0.024 | 1.031 | 0.827 | 0.345 |
| TACN | 96 | 1.348 | 0.851 | 0.321 | 0.381 | 0.496 | 0.713 | 0.574 | 0.599 | -0.230 | 0.703 | 0.640 | 0.186 | 0.084 | 0.230 | 0.100 |
| | 192 | 1.396 | 0.895 | 0.445 | 0.669 | 0.657 | 0.430 | 7.302 | 2.342 | 0.337 | 0.717 | 0.654 | 0.147 | 0.166 | 0.333 | 0.234 |
| | 336 | 1.301 | 0.838 | 0.447 | 0.839 | 0.742 | 0.374 | 5.918 | 1.972 | 0.151 | **0.737** | 0.662 | 0.107 | 0.313 | 0.459 | 0.107 |
| | 720 | 1.483 | 0.878 | 0.414 | - | - | - | - | - | - | 0.761 | 0.668 | 0.056 | 3.910 | 1.718 | 0.015 |
| IMV-LSTM | 96 | 1.355 | 0.928 | 0.520 | 0.616 | 0.630 | 0.903 | 18.11 | 4.056 | **0.376** | 0.672 | 0.689 | 0.246 | 0.759 | 0.761 | 0.260 |
| | 192 | 1.286 | 0.918 | 0.473 | 0.834 | 0.756 | 0.808 | 16.36 | 3.778 | 0.312 | 0.782 | 0.712 | 0.139 | 0.986 | 0.859 | **0.374** |
| | 336 | 1.202 | 0.831 | 0.427 | 1.178 | 0.890 | 0.754 | 24.07 | 4.604 | **0.430** | 0.741 | 0.661 | 0.113 | 1.267 | 0.950 | **0.401** |
| | 720 | 1.212 | 0.806 | 0.323 | - | - | - | - | - | - | 0.754 | 0.678 | 0.062 | 1.676 | 1.105 | 0.674 |
| MTEX | 96 | 1.308 | 0.870 | 0.467 | 0.597 | 0.624 | 0.897 | 13.20 | 3.401 | 0.090 | 0.725 | 0.671 | 0.009 | 0.579 | 0.539 | -0.030 |
| | 192 | 1.449 | 0.950 | 0.460 | 1.416 | 0.950 | 0.603 | 10.00 | 2.544 | 0.163 | 0.721 | 0.666 | 0.036 | 2.858 | 1.332 | -0.070 |
| | 336 | 1.412 | 0.940 | 0.457 | 1.712 | 1.037 | 0.712 | 20.56 | 4.256 | 0.054 | 0.739 | 0.673 | 0.027 | 1.893 | 1.040 | -0.010 |
| | 720 | 1.876 | 0.974 | 0.065 | - | - | - | - | - | - | 0.758 | 0.670 | 0.002 | 3.301 | 1.678 | 0.327 |
| XCM | 96 | 1.312 | 0.867 | 0.362 | 0.705 | 0.657 | 0.803 | 6.873 | 1.841 | 0.150 | 0.691 | 0.628 | 0.230 | 0.315 | 0.418 | 0.091 |
| | 192 | 1.308 | 0.876 | 0.485 | 1.316 | 0.945 | 0.728 | 8.407 | 2.398 | -0.160 | 0.724 | 0.652 | 0.140 | 1.276 | 0.981 | 0.093 |
| | 336 | 1.309 | 0.873 | 0.388 | 1.430 | 0.964 | 0.589 | 19.58 | 3.860 | 0.054 | 0.748 | 0.672 | 0.091 | 1.501 | 1.096 | 0.043 |
| | 720 | 1.748 | 0.975 | 0.166 | - | - | - | - | - | - | 0.758 | 0.676 | 0.026 | 3.832 | 1.699 | 0.589 |

TABLE VI
RESULTS OF COMPARATIVE EXPERIMENTS ON LONG-TERM TIME SERIES FORECASTING

| Dataset | | Weather | | | ETTh1 | | | ETTm1 | | | ETTh2 | | | ETTm2 | | |
|---|---|---|---|---|---|---|---|---|---|---|---|---|---|---|---|---|
| Metrics | | MSE | MAE | COR | MSE | MAE | COR | MSE | MAE | COR | MSE | MAE | COR | MSE | MAE | COR |
| Ours | 96 | 0.171 | 0.307 | 0.473 | 0.117 | 0.258 | 0.202 | 0.047 | 0.162 | **0.284** | 0.143 | 0.292 | 0.680 | **0.074** | **0.196** | **0.725** |
| | 192 | 0.231 | 0.360 | **0.476** | 0.161 | 0.315 | 0.156 | 0.093 | 0.228 | 0.210 | **0.166** | **0.319** | 0.597 | **0.109** | **0.247** | 0.672 |
| | 336 | 0.340 | 0.447 | **0.450** | 0.181 | 0.333 | 0.176 | 0.139 | 0.285 | 0.205 | **0.213** | **0.369** | 0.549 | **0.140** | 0.288 | **0.641** |
| | 720 | 0.430 | 0.470 | 0.320 | 0.137 | 0.296 | **0.259** | 0.216 | 0.377 | 0.172 | **0.205** | **0.368** | 0.509 | **0.160** | **0.314** | 0.588 |
| iTransformer | 96 | 0.121 | 0.247 | 0.469 | 0.058 | 0.185 | 0.261 | **0.029** | **0.127** | 0.281 | **0.133** | **0.280** | **0.698** | **0.074** | 0.197 | 0.712 |
| | 192 | 0.150 | **0.281** | 0.458 | 0.079 | **0.215** | **0.224** | **0.044** | **0.160** | 0.244 | 0.181 | 0.333 | **0.647** | 0.110 | **0.247** | **0.676** |
| | 336 | **0.217** | **0.338** | 0.413 | **0.093** | **0.237** | **0.221** | **0.059** | **0.187** | **0.208** | 0.231 | 0.381 | **0.613** | 0.141 | **0.287** | 0.634 |
| | 720 | **0.318** | **0.415** | **0.439** | **0.090** | **0.237** | 0.223 | **0.078** | **0.215** | 0.219 | 0.252 | 0.403 | **0.558** | 0.181 | 0.331 | **0.623** |
| NLinear | 96 | 0.137 | 0.263 | 0.373 | **0.057** | **0.182** | 0.262 | 0.031 | 0.131 | 0.235 | 0.134 | 0.281 | 0.658 | 0.083 | 0.211 | 0.675 |
| | 192 | 0.163 | 0.291 | 0.372 | **0.078** | **0.215** | 0.214 | 0.046 | 0.162 | 0.207 | 0.185 | 0.335 | 0.595 | 0.115 | 0.253 | 0.628 |
| | 336 | 0.231 | 0.347 | 0.341 | 0.096 | 0.243 | 0.196 | 0.060 | 0.188 | 0.172 | 0.241 | 0.389 | 0.557 | 0.146 | 0.290 | 0.580 |
| | 720 | 0.338 | 0.429 | 0.404 | 0.106 | 0.257 | 0.157 | 0.082 | 0.220 | 0.159 | 0.305 | 0.447 | 0.485 | 0.188 | 0.335 | 0.557 |
| TFT | 96 | 0.109 | **0.238** | **0.498** | 0.092 | 0.232 | 0.191 | 0.047 | 0.163 | 0.217 | 0.295 | 0.434 | 0.593 | 0.110 | 0.242 | 0.631 |
| | 192 | 0.160 | 0.304 | 0.475 | 0.101 | 0.246 | 0.102 | 0.089 | 0.232 | 0.059 | 0.310 | 0.455 | 0.529 | 0.266 | 0.399 | 0.579 |
| | 336 | 0.304 | 0.422 | 0.324 | 0.150 | 0.306 | 0.124 | 0.140 | 0.289 | 0.010 | 0.384 | 0.506 | 0.362 | 0.699 | 0.670 | 0.220 |
| | 720 | 0.422 | 0.500 | 0.154 | 0.344 | 0.484 | -0.030 | 0.113 | 0.259 | 0.028 | 0.920 | 0.823 | 0.333 | 0.393 | 0.520 | 0.027 |
| TACN | 96 | 0.128 | 0.267 | 0.402 | 0.238 | 0.413 | 0.065 | 0.054 | 0.177 | 0.215 | 0.384 | 0.503 | 0.192 | 0.111 | 0.255 | 0.568 |
| | 192 | 0.169 | 0.318 | 0.458 | 0.297 | 0.468 | 0.081 | 0.094 | 0.233 | 0.140 | 1.173 | 0.853 | 0.131 | 0.213 | 0.360 | 0.432 |
| | 336 | 0.269 | 0.404 | 0.394 | 0.222 | 0.389 | 0.177 | 0.203 | 0.376 | 0.039 | 1.251 | 0.946 | 0.231 | 0.632 | 0.657 | 0.184 |
| | 720 | 0.371 | 0.481 | 0.363 | 0.361 | 0.513 | -0.050 | 0.334 | 0.493 | 0.022 | 0.974 | 0.861 | -0.020 | 1.960 | 1.156 | 0.158 |
| IMV-LSTM | 96 | 0.173 | 0.312 | 0.440 | 0.447 | 0.569 | 0.245 | 0.165 | 0.326 | 0.224 | 1.374 | 0.996 | 0.274 | 0.280 | 0.415 | 0.536 |
| | 192 | 0.240 | 0.368 | 0.453 | 0.768 | 0.787 | 0.154 | 0.315 | 0.474 | 0.188 | 1.785 | 1.170 | 0.194 | 0.262 | 0.395 | 0.444 |
| | 336 | 0.351 | 0.455 | 0.402 | 0.618 | 0.691 | 0.175 | 0.511 | 0.633 | 0.191 | 2.170 | 1.323 | 0.255 | 0.673 | 0.675 | 0.372 |
| | 720 | 0.444 | 0.510 | 0.318 | 1.190 | 1.013 | 0.244 | 1.194 | 1.027 | 0.154 | 1.707 | 1.157 | 0.285 | 2.051 | 1.256 | 0.194 |
| MTEX | 96 | **0.107** | 0.252 | 0.464 | 0.182 | 0.347 | **0.331** | 0.331 | 0.375 | 0.275 | 0.551 | 0.579 | 0.367 | 0.448 | 0.478 | 0.431 |
| | 192 | **0.146** | 0.293 | 0.468 | 0.165 | 0.325 | 0.108 | 0.098 | 0.234 | **0.321** | 0.495 | 0.559 | 0.472 | 0.350 | 0.440 | 0.517 |
| | 336 | 0.226 | 0.370 | 0.442 | 0.384 | 0.467 | 0.102 | 0.483 | 0.529 | 0.200 | 0.362 | 0.483 | 0.546 | 0.457 | 0.523 | 0.433 |
| | 720 | 0.476 | 0.538 | 0.316 | 0.214 | 0.376 | 0.029 | 0.420 | 0.500 | 0.165 | 0.448 | 0.529 | 0.335 | 0.628 | 0.627 | 0.263 |
| XCM | 96 | 0.109 | 0.246 | 0.464 | 0.146 | 0.307 | 0.197 | 0.055 | 0.177 | 0.281 | 0.331 | 0.459 | 0.570 | 0.144 | 0.287 | 0.704 |
| | 192 | 0.158 | 0.305 | 0.467 | 0.163 | 0.313 | 0.141 | 0.094 | 0.237 | 0.207 | 0.399 | 0.509 | 0.540 | 0.216 | 0.356 | 0.644 |
| | 336 | 0.243 | 0.376 | 0.394 | 0.145 | 0.335 | 0.154 | 0.140 | 0.295 | 0.144 | 0.375 | 0.496 | 0.477 | 0.304 | 0.431 | 0.584 |
| | 720 | 0.381 | 0.488 | 0.391 | 0.231 | 0.386 | 0.133 | 0.177 | 0.335 | **0.284** | 0.384 | 0.500 | 0.434 | 0.414 | 0.510 | 0.506 |

TABLE VII
RESULTS OF COMPARATIVE EXPERIMENTS ON SHORT-TERM TIME SERIES FORECASTING

| Dataset | | PAQ | | | SML | | | Volatility | | | PM2.5 | | | Exchange | | |
|---|---|---|---|---|---|---|---|---|---|---|---|---|---|---|---|---|
| Metrics | | MSE | MAE | COR | MSE | MAE | COR | MSE | MAE | COR | MSE | MAE | COR | MSE | MAE | COR |
| Ours | 3 | 0.455 | 0.452 | 0.473 | **0.000** | 0.015 | 0.944 | 0.029 | 0.124 | 0.018 | **0.105** | **0.197** | **0.205** | **0.003** | **0.039** | **0.066** |
| | 6 | 0.591 | **0.519** | 0.637 | **0.001** | **0.025** | 0.952 | 0.034 | 0.129 | 0.059 | **0.190** | **0.276** | **0.236** | **0.005** | 0.054 | 0.088 |
| | 12 | **0.684** | **0.562** | 0.691 | 0.007 | 0.052 | 0.953 | 0.072 | 0.193 | 0.030 | **0.313** | **0.374** | **0.291** | 0.010 | 0.077 | 0.102 |
| | 36 | **0.880** | **0.643** | **0.702** | 0.090 | 0.196 | 0.919 | 0.140 | 0.277 | **0.145** | **0.549** | 0.541 | **0.281** | 0.026 | 0.126 | **0.189** |
| iTransformer | 3 | **0.423** | **0.449** | **0.556** | 0.001 | 0.017 | **0.950** | 0.020 | 0.098 | 0.041 | 0.118 | 0.207 | 0.131 | **0.003** | 0.040 | 0.020 |
| | 6 | **0.564** | 0.525 | **0.669** | 0.002 | 0.026 | **0.955** | 0.035 | 0.129 | 0.040 | 0.209 | 0.283 | 0.164 | **0.005** | **0.052** | 0.029 |
| | 12 | 0.686 | 0.582 | **0.700** | **0.006** | **0.048** | **0.956** | 0.060 | 0.172 | 0.064 | 0.352 | 0.382 | 0.214 | **0.009** | **0.070** | 0.027 |
| | 36 | 0.919 | 0.669 | 0.663 | **0.067** | **0.173** | **0.946** | 0.133 | 0.272 | 0.089 | 0.662 | 0.557 | 0.213 | **0.024** | **0.119** | 0.010 |
| NLinear | 3 | 0.516 | 0.494 | 0.502 | **0.000** | **0.013** | 0.922 | **0.019** | **0.095** | -0.010 | 0.122 | 0.206 | 0.051 | **0.003** | 0.042 | 0.034 |
| | 6 | 0.683 | 0.575 | 0.634 | **0.001** | 0.026 | 0.906 | **0.032** | 0.127 | -0.020 | 0.229 | 0.290 | 0.095 | **0.005** | 0.054 | -0.010 |
| | 12 | 0.819 | 0.628 | 0.653 | 0.010 | 0.064 | 0.883 | **0.054** | **0.167** | -0.010 | 0.400 | 0.400 | 0.147 | 0.010 | 0.072 | -0.020 |
| | 36 | 1.053 | 0.718 | 0.611 | 0.156 | 0.270 | 0.844 | **0.128** | **0.265** | 0.074 | 0.727 | 0.580 | 0.157 | 0.026 | 0.123 | -0.040 |
| TFT | 3 | 0.471 | 0.467 | 0.503 | 0.001 | 0.019 | 0.898 | 0.026 | 0.113 | -0.020 | 0.113 | 0.203 | 0.140 | **0.003** | 0.040 | -0.040 |
| | 6 | 0.718 | 0.592 | 0.668 | 0.009 | 0.060 | 0.897 | 0.046 | 0.149 | 0.032 | 0.216 | 0.287 | 0.183 | **0.005** | 0.055 | -0.030 |
| | 12 | 0.802 | 0.621 | 0.677 | 0.016 | 0.090 | 0.899 | 0.059 | 0.178 | 0.000 | 0.360 | 0.393 | 0.231 | **0.009** | 0.071 | 0.029 |
| | 36 | 1.415 | 0.863 | 0.661 | 0.305 | 0.386 | 0.768 | 0.176 | 0.321 | 0.007 | 0.609 | 0.551 | 0.186 | 0.031 | 0.138 | -0.100 |
| TACN | 3 | 0.514 | 0.491 | 0.364 | 0.002 | 0.035 | 0.418 | 0.020 | 0.097 | -0.020 | 0.123 | 0.215 | 0.069 | 0.004 | 0.043 | -0.010 |
| | 6 | 0.696 | 0.586 | 0.437 | 0.008 | 0.065 | 0.391 | 0.033 | **0.123** | -0.040 | 0.220 | 0.299 | 0.141 | 0.006 | 0.057 | -0.030 |
| | 12 | 0.839 | 0.651 | 0.463 | 0.025 | 0.104 | 0.712 | 0.058 | 0.168 | 0.021 | 0.352 | 0.398 | 0.231 | 0.021 | 0.090 | 0.000 |
| | 36 | 1.098 | 0.747 | 0.522 | 0.177 | 0.294 | 0.692 | 0.132 | 0.269 | 0.104 | 0.572 | 0.547 | 0.202 | 0.027 | 0.127 | -0.010 |
| IMV-LSTM | 3 | 0.557 | 0.526 | 0.529 | 0.030 | 0.118 | 0.914 | 8.606 | 2.626 | **0.101** | **0.105** | 0.199 | 0.193 | 0.245 | 0.383 | 0.060 |
| | 6 | 0.790 | 0.623 | 0.610 | 0.057 | 0.158 | 0.909 | 11.51 | 3.130 | **0.135** | 0.192 | 0.284 | **0.236** | 0.287 | 0.418 | **0.101** |
| | 12 | 0.978 | 0.705 | 0.608 | 0.072 | 0.194 | 0.903 | 13.01 | 3.353 | **0.099** | **0.313** | 0.378 | 0.286 | 0.390 | 0.519 | **0.126** |
| | 36 | 1.311 | 0.840 | 0.579 | 0.231 | 0.360 | 0.916 | 17.01 | 3.903 | 0.127 | 0.551 | **0.515** | 0.279 | 0.723 | 0.733 | 0.187 |
| MTEX | 3 | 0.724 | 0.563 | 0.370 | 0.006 | 0.060 | 0.890 | 10.95 | 2.647 | -0.060 | 0.286 | 0.338 | 0.113 | 0.240 | 0.298 | 0.031 |
| | 6 | 0.909 | 0.654 | 0.512 | 0.014 | 0.082 | 0.869 | 11.71 | 2.911 | -0.060 | 0.265 | 0.336 | 0.187 | 0.305 | 0.357 | 0.072 |
| | 12 | 1.034 | 0.713 | 0.507 | 0.423 | 0.346 | 0.710 | 8.556 | 2.284 | -0.030 | 0.368 | 0.416 | 0.215 | 0.301 | 0.342 | 0.081 |
| | 36 | 1.270 | 0.823 | 0.444 | 0.162 | 0.305 | 0.869 | 9.513 | 2.340 | -0.210 | 0.606 | 0.585 | 0.177 | 0.646 | 0.562 | 0.007 |
| XCM | 3 | 0.687 | 0.603 | 0.403 | 0.019 | 0.105 | 0.508 | 0.490 | 0.592 | 0.070 | 0.119 | 0.227 | 0.145 | 0.237 | 0.367 | 0.006 |
| | 6 | 0.909 | 0.711 | 0.465 | 0.050 | 0.166 | 0.662 | 0.390 | 0.669 | 0.050 | 0.203 | 0.301 | 0.185 | 0.236 | 0.312 | 0.010 |
| | 12 | 1.045 | 0.750 | 0.492 | 0.051 | 0.163 | 0.811 | 0.382 | 0.489 | 0.009 | 0.341 | 0.410 | 0.236 | 0.155 | 0.264 | 0.022 |
| | 36 | 1.232 | 0.830 | 0.531 | 0.187 | 0.326 | 0.866 | 0.690 | 0.653 | -0.190 | 0.561 | 0.556 | 0.270 | 0.097 | 0.239 | 0.046 |

TABLE VIII

RESULTS OF COMPARATIVE EXPERIMENTS ON SHORT-TERM TIME SERIES FORECASTING

| Dataset | | Weather | | | ETTh1 | | | ETTm1 | | | ETTh2 | | | ETTm2 | | |
|---|---|---|---|---|---|---|---|---|---|---|---|---|---|---|---|---|
| Metrics | | MSE | MAE | COR | MSE | MAE | COR | MSE | MAE | COR | MSE | MAE | COR | MSE | MAE | COR |
| Ours | 3 | 0.002 | 0.027 | **0.306** | 0.010 | 0.076 | 0.143 | 0.003 | 0.037 | **0.096** | **0.010** | **0.067** | **0.760** | **0.001** | **0.019** | **0.798** |
| | 6 | 0.004 | 0.041 | **0.353** | 0.019 | 0.102 | 0.211 | 0.005 | 0.052 | **0.114** | **0.023** | **0.102** | **0.778** | **0.002** | **0.031** | 0.721 |
| | 12 | 0.014 | 0.087 | 0.355 | 0.029 | 0.128 | 0.303 | 0.008 | 0.063 | **0.151** | **0.045** | **0.150** | **0.802** | **0.008** | **0.055** | **0.711** |
| | 36 | 0.091 | 0.208 | 0.308 | 0.058 | 0.185 | 0.286 | 0.025 | 0.118 | **0.258** | 0.094 | 0.229 | 0.740 | **0.047** | **0.138** | **0.743** |
| iTransformer | 3 | **0.001** | 0.024 | 0.304 | **0.007** | **0.061** | **0.201** | **0.002** | 0.033 | 0.054 | 0.011 | 0.070 | 0.740 | **0.001** | **0.019** | 0.771 |
| | 6 | **0.003** | **0.037** | 0.341 | **0.012** | **0.079** | **0.271** | **0.004** | **0.042** | 0.092 | 0.029 | 0.113 | 0.744 | **0.002** | **0.031** | **0.727** |
| | 12 | 0.009 | **0.062** | **0.356** | **0.019** | **0.104** | **0.327** | **0.006** | **0.056** | 0.130 | 0.055 | 0.164 | 0.796 | **0.008** | 0.056 | 0.707 |
| | 36 | 0.057 | 0.152 | **0.339** | 0.037 | 0.147 | 0.292 | **0.016** | **0.093** | 0.206 | **0.089** | **0.223** | **0.744** | 0.053 | 0.144 | 0.707 |
| NLinear | 3 | **0.001** | 0.024 | 0.258 | **0.007** | 0.062 | 0.150 | **0.002** | **0.032** | 0.019 | 0.013 | 0.074 | 0.693 | 0.002 | 0.026 | 0.636 |
| | 6 | **0.003** | **0.037** | 0.300 | **0.012** | 0.081 | 0.224 | **0.004** | **0.042** | 0.028 | 0.033 | 0.118 | 0.697 | 0.004 | 0.041 | 0.567 |
| | 12 | 0.010 | 0.063 | 0.304 | 0.020 | **0.104** | 0.317 | 0.007 | 0.057 | 0.054 | 0.058 | 0.169 | 0.755 | 0.014 | 0.075 | 0.403 |
| | 36 | 0.061 | 0.155 | 0.214 | **0.036** | **0.144** | 0.314 | 0.017 | 0.095 | 0.163 | 0.090 | **0.223** | 0.719 | 0.096 | 0.204 | 0.488 |
| TFT | 3 | **0.001** | **0.023** | **0.306** | 0.009 | 0.070 | 0.167 | **0.002** | 0.034 | 0.055 | 0.016 | 0.074 | 0.720 | **0.001** | **0.019** | 0.728 |
| | 6 | 0.004 | 0.048 | 0.218 | 0.016 | 0.095 | 0.228 | **0.004** | 0.045 | 0.070 | 0.035 | 0.116 | 0.766 | 0.003 | 0.032 | 0.663 |
| | 12 | 0.012 | 0.069 | 0.323 | 0.030 | 0.130 | 0.296 | 0.008 | 0.060 | 0.142 | 0.072 | 0.188 | 0.783 | 0.012 | 0.066 | 0.589 |
| | 36 | 0.060 | 0.158 | 0.143 | 0.083 | 0.223 | **0.335** | 0.025 | 0.118 | 0.145 | 0.193 | 0.344 | 0.712 | 0.082 | 0.192 | 0.538 |
| TACN | 3 | 0.002 | 0.029 | 0.165 | 0.010 | 0.078 | 0.168 | 0.003 | 0.037 | 0.008 | 0.018 | 0.098 | 0.535 | 0.002 | 0.030 | 0.388 |
| | 6 | 0.005 | 0.045 | 0.210 | 0.020 | 0.102 | 0.190 | 0.005 | 0.053 | 0.050 | 0.035 | 0.131 | 0.584 | 0.004 | 0.043 | 0.350 |
| | 12 | 0.009 | 0.065 | 0.193 | 0.031 | 0.138 | 0.155 | 0.009 | 0.068 | 0.093 | 0.065 | 0.187 | 0.700 | 0.030 | 0.111 | 0.316 |
| | 36 | 0.056 | 0.166 | 0.228 | 0.134 | 0.305 | 0.043 | 0.026 | 0.119 | 0.208 | 0.134 | 0.285 | 0.593 | 0.110 | 0.228 | 0.421 |
| IMV-LSTM | 3 | 0.003 | 0.027 | 0.305 | 0.015 | 0.092 | 0.138 | 0.005 | 0.052 | 0.057 | 0.017 | 0.091 | 0.674 | 0.003 | 0.027 | 0.730 |
| | 6 | 0.005 | 0.045 | 0.321 | 0.030 | 0.132 | 0.151 | 0.013 | 0.085 | 0.057 | 0.062 | 0.177 | 0.604 | 0.005 | 0.049 | 0.617 |
| | 12 | 0.012 | 0.076 | 0.314 | 0.077 | 0.215 | 0.258 | 0.023 | 0.118 | 0.092 | 0.104 | 0.238 | 0.631 | 0.016 | 0.086 | 0.503 |
| | 36 | 0.092 | 0.214 | 0.292 | 0.186 | 0.342 | 0.226 | 0.087 | 0.227 | 0.219 | 0.328 | 0.446 | 0.529 | 0.144 | 0.273 | 0.491 |
| MTEX | 3 | 0.004 | 0.036 | 0.004 | 0.013 | 0.080 | 0.011 | 0.005 | 0.049 | 0.018 | 0.019 | 0.099 | 0.530 | 0.004 | 0.048 | 0.145 |
| | 6 | 0.005 | 0.049 | 0.021 | 0.021 | 0.104 | 0.165 | 0.007 | 0.057 | 0.012 | 0.190 | 0.251 | 0.593 | 0.008 | 0.064 | 0.294 |
| | 12 | 0.015 | 0.095 | 0.252 | 0.036 | 0.142 | 0.300 | 0.012 | 0.076 | 0.016 | 0.092 | 0.218 | 0.754 | 0.199 | 0.228 | 0.316 |
| | 36 | **0.042** | **0.146** | 0.308 | 0.290 | 0.357 | 0.252 | 0.026 | 0.119 | 0.221 | 0.306 | 0.407 | 0.579 | 0.414 | 0.431 | 0.310 |
| XCM | 3 | 0.003 | 0.027 | 0.245 | 0.023 | 0.116 | 0.134 | 0.005 | 0.052 | 0.090 | 0.025 | 0.117 | 0.593 | 0.003 | 0.035 | 0.598 |
| | 6 | 0.005 | 0.046 | 0.315 | 0.031 | 0.133 | 0.204 | 0.008 | 0.065 | 0.101 | 0.062 | 0.186 | 0.656 | 0.006 | 0.056 | 0.504 |
| | 12 | **0.008** | 0.064 | 0.320 | 0.044 | 0.160 | 0.250 | 0.014 | 0.085 | 0.149 | 0.109 | 0.246 | 0.692 | 0.018 | 0.094 | 0.556 |
| | 36 | 0.047 | 0.156 | 0.305 | 0.076 | 0.213 | 0.273 | 0.031 | 0.129 | 0.236 | 0.203 | 0.352 | 0.641 | 0.089 | 0.216 | 0.619 |

TABLE IX

OVERALL AVERAGE RANKING OF FORECASTING RESULTS

| Models | Long | Short | **AVG** |
|---|---|---|---|
| Ours | **2.324** | 2.166 | **2.245** |
| iTransformer | 2.368 | **2.141** | 2.254 |
| NLinear | 3.526 | 3.475 | 3.500 |
| TFT | 4.947 | 3.808 | 4.377 |
| TACN | 5.456 | 4.800 | 5.128 |
| IMV-LSTM | 5.894 | 5.058 | 5.476 |
| MTEX | 5.798 | 6.091 | 5.944 |
| XCM | 4.929 | 5.208 | 5.068 |

overall.

DLFormer shows consistently strong and stable performance in both long- and short-term forecasting, achieving top overall rankings. Despite being MDE-based, it delivers state-of-the-art results across various datasets. This highlights the effectiveness of the TVAL strategy, which prevents dimensional mixing in Transformer models and improves forecasting accuracy by capturing complex time and variable-wise dependencies.

**Quantitative Explainability** Performance Tables X and XI present the robustness evaluation results for the explanations generated by each explainable time-series model across different datasets. Each value reflects the outcomes from applying 10 independently trained models to the complete test set. The robustness of VI scores, extracted from multi-dimensional explanations, is assessed using percentage- and rank-based metrics. For DLFormer, we used the $\hat{\alpha}_{Var}$ described in Section III-3. The best results are highlighted in bold red and second-best results are blue underline. In Table X, MTEX demonstrates superior robustness in percentage-based VI, achieving the best results across all evaluation metrics. DLFormer mostly ranks second, exhibiting consistently high robustness.

In contrast, Table XI, which evaluates the consistency of variable rankings, shows that DLFormer achieves the best robustness in most cases. XCM ranks second, followed by MTEX, TFT, and IMV-LSTM in that order. From a quantitative perspective, the average variability of VI percentages across individually trained DLFormer models is $2.123\%$, indicating high stability. In terms of ranking consistency, the average COR between the individually trained DLFormer models reaches $67.6\%$, demonstrating remarkable robustness compared to most other models. Table XII summarizes the overall ranking for the quantitative evaluation of explainability, averaging rankings of each metric across all datasets. The results confirm that DLFormer achieves the best comprehensive performance in terms of explanation robustness. Thus, based on forecasting performance comparisons across various settings and quantitative explainability, DLFormer clearly outperforms existing explainable time-series methods in both aspects.

Fig. 4 illustrates the trade-off between forecasting performance and explainability. It compares where models stand in terms of both, with lower ranks being better. MDE-based models like MTEX show moderate forecasting but strong explainability. In contrast, UDE-based models like TACN are only evaluated on forecasting due to limited explainability. DLFormer stands out by achieving both high accuracy and strong explainability, effectively reducing the usual trade-off.

TABLE X
ROBUSTNESS EVALUATION RESULTS FOR EXPLANATIONS: PERCENTAGE-BASED

| Models | Ours | | TFT | | IMV-LSTM | | MTEX | | XCM | |
|---|---|---|---|---|---|---|---|---|---|---|
| Metrics | STD | CV | STD | CV | STD | CV | STD | CV | STD | CV |
| PAQ | 1.624 | 0.111 | 7.351 | 0.501 | 3.473 | 0.242 | **0.215** | **0.015** | 1.855 | 0.137 |
| SML | 0.685 | 0.110 | 3.648 | 0.564 | 2.769 | 0.403 | **0.027** | **0.004** | 2.328 | 0.446 |
| Volatility | 3.085 | 0.240 | 7.164 | 0.562 | 3.196 | 0.252 | **0.320** | **0.025** | 4.615 | 0.396 |
| PM2.5 | 2.047 | 0.149 | 5.480 | 0.445 | 2.645 | 0.194 | **0.046** | **0.003** | 4.320 | 0.317 |
| Exchange | 3.765 | 0.345 | 6.243 | 0.512 | 4.377 | 0.361 | **0.174** | **0.013** | 2.541 | 0.202 |
| Weather | 1.739 | 0.231 | 3.071 | 0.462 | 1.509 | 0.194 | **0.016** | **0.002** | 1.150 | 0.160 |
| ETTh1 | 2.454 | 0.164 | 8.089 | 0.590 | 5.648 | 0.410 | **0.062** | **0.004** | 2.575 | 0.179 |
| ETTm1 | 2.719 | 0.187 | 5.134 | 0.431 | 5.254 | 0.371 | **0.033** | **0.002** | 3.017 | 0.208 |
| ETTh2 | 1.891 | 0.142 | 7.907 | 0.591 | 5.303 | 0.403 | **0.537** | **0.037** | 2.686 | 0.203 |
| ETTm2 | 2.155 | 0.179 | 6.129 | 0.506 | 6.000 | 0.508 | **0.112** | **0.078** | 2.175 | 0.186 |

TABLE XI
ROBUSTNESS EVALUATION RESULTS FOR EXPLANATIONS: RANK-BASED

| Models | Ours | | TFT | | IMV-LSTM | | MTEX | | XCM | |
|---|---|---|---|---|---|---|---|---|---|---|
| Metrics | TAU | COR | TAU | COR | TAU | COR | TAU | COR | TAU | COR |
| PAQ | **0.765** | **0.885** | 0.121 | 0.158 | 0.206 | 0.252 | 0.299 | 0.368 | 0.348 | 0.428 |
| SML | **0.720** | **0.857** | 0.118 | 0.168 | 0.269 | 0.369 | 0.269 | 0.351 | 0.322 | 0.372 |
| Volatility | 0.374 | **0.480** | 0.195 | 0.251 | 0.126 | 0.165 | 0.157 | 0.200 | **0.428** | 0.429 |
| PM2.5 | **0.720** | **0.821** | 0.316 | 0.395 | 0.303 | 0.404 | 0.482 | 0.561 | 0.301 | 0.340 |
| Exchange | **0.625** | **0.777** | 0.215 | 0.278 | 0.041 | 0.075 | 0.238 | 0.234 | 0.169 | 0.186 |
| Weather | 0.479 | 0.616 | 0.436 | 0.567 | 0.046 | 0.055 | 0.118 | 0.134 | **0.557** | **0.635** |
| ETTh1 | **0.722** | **0.813** | 0.113 | 0.164 | 0.180 | 0.252 | 0.610 | 0.666 | 0.458 | 0.573 |
| ETTm1 | 0.519 | 0.668 | 0.377 | 0.492 | 0.297 | 0.373 | **0.731** | **0.840** | 0.238 | 0.326 |
| ETTh2 | **0.657** | **0.769** | 0.066 | 0.103 | 0.266 | 0.356 | 0.253 | 0.285 | 0.291 | 0.319 |
| ETTm2 | **0.598** | **0.707** | 0.299 | 0.363 | 0.243 | 0.330 | 0.582 | 0.652 | 0.208 | 0.230 |

TABLE XII
ROBUSTNESS EVALUATION RESULTS FOR EXPLANATIONS: RANK-BASED

| Models | Percentage | Rank | **AVG** |
|---|---|---|---|
| Ours | 2.30 | **1.25** | **1.77** |
| TFT | 4.90 | 3.75 | 4.32 |
| IMV-LSTM | 3.75 | 4.00 | 3.87 |
| MTEX | **1.00** | 2.80 | 1.90 |
| XCM | 3.05 | 3.10 | 3.07 |

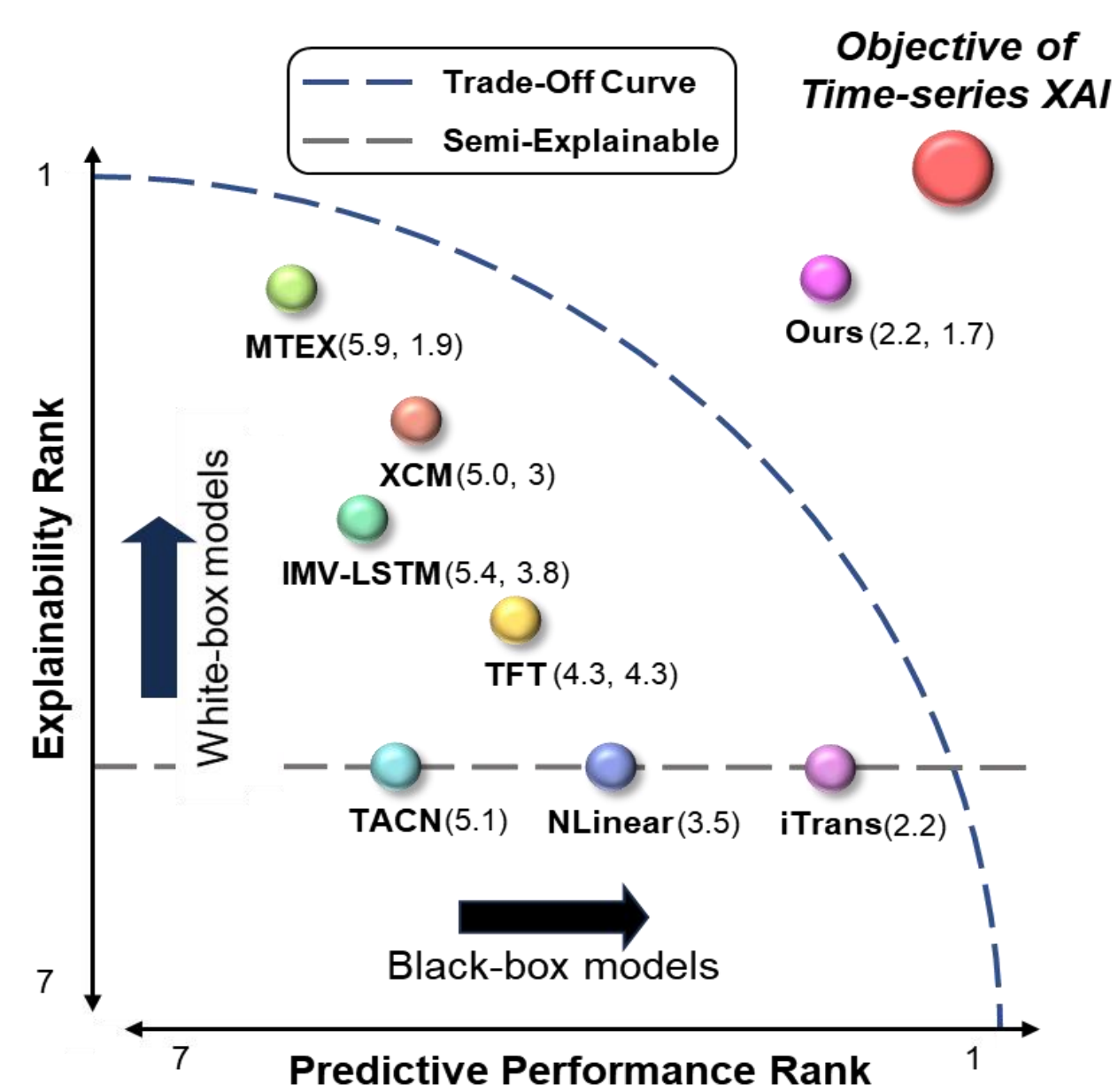

**Fig. 4.** Performance-explainability trade-off for time-series forecasting models.

**Qualitative Evaluation of Explainability** For the qualitative explainability evaluation, we utilized two datasets related to environmental pollution, PAQ and PM2.5, which are well-studied in the literature. These datasets facilitate qualitative comparisons based on domain knowledge. Fig. 5 presents the visualization of the DLFormer TVA attention maps for the two datasets. Fig. 6 visualizes feature importance extracted from these two datasets using five different models, analyzing from TI and VI perspectives. A detailed explanation of these results follows.

**a) Port Air Quality Dataset Analysis**: Air quality in port environments typically exhibits a 24-hour periodicity due to ongoing port operations [39], [40]. Ship arrivals, departures, and vehicular activities predominantly occur during the diurnal period, generating concentrated air pollution. Emissions from these activities interact and have a prolonged impact on air quality over subsequent hours [41]. As shown in Fig. 5, the TVA attention map for the PAQ dataset reveals that DLFormer consistently emphasizes the NOx predictor values from 24 hours prior at each prediction step.

This observation aligns well with domain knowledge regarding the daily periodicity of port emissions. Such periodicity is observed across all variables, indicating strong correlations driven by the cyclical nature of air quality in port operations.

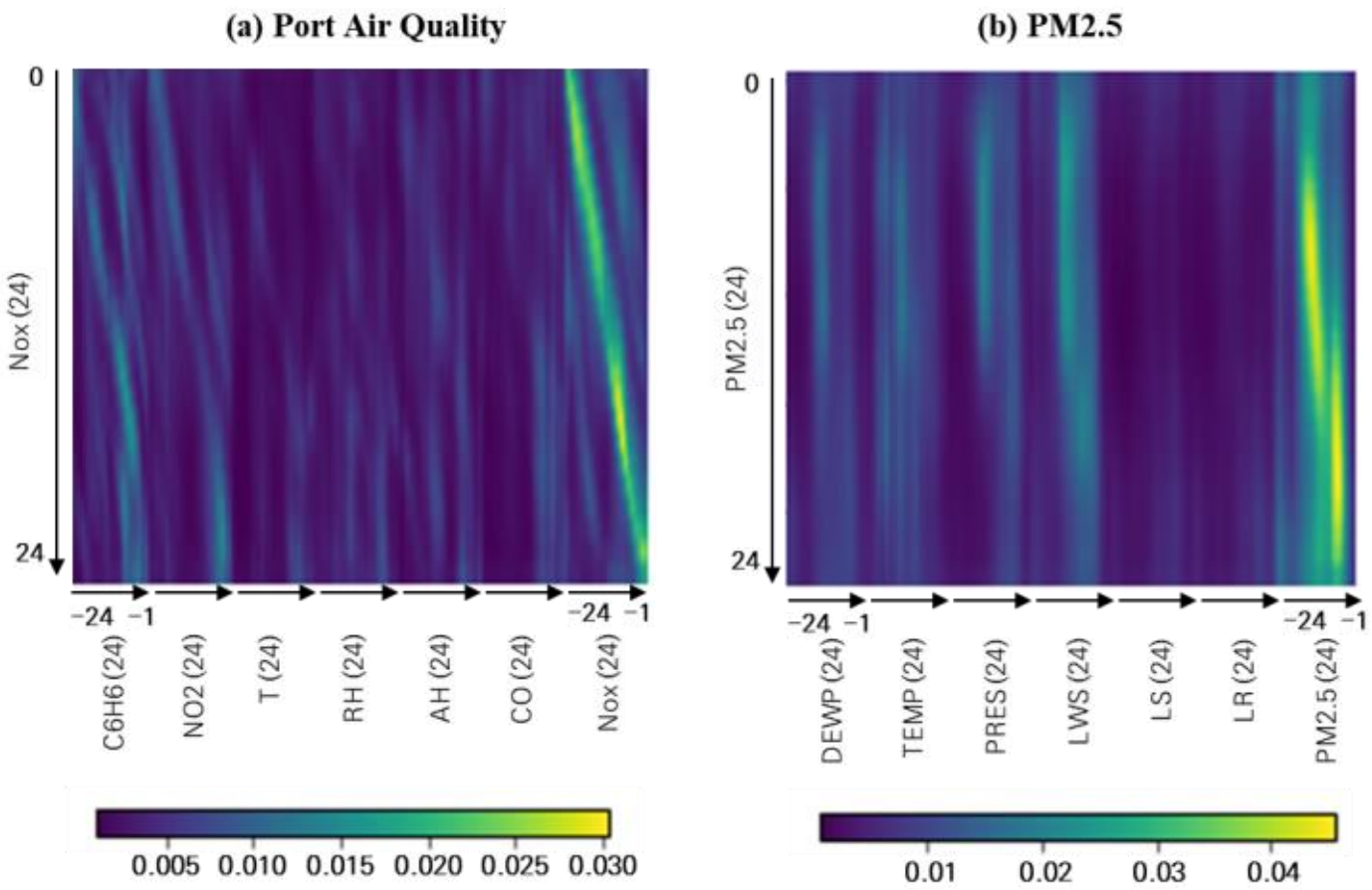

**Fig. 5.** Visualization of the time-variable-aware attention map of DLFormer for (a) Port Air Quality and (b) PM2.5 datasets.

In Fig. 6, the DLFormer results for the PAQ dataset, aggregated over forecasting horizons, reflect the general influence of past time points. While some noise exists, the model emphasizes time points from 10 to 13 hours prior, suggesting that daytime emissions persistently impact air quality later in the day. For VI, NOx exhibits the highest contribution due to autoregressive characteristics. NO2, a component of NOx, shows the second-highest importance. Additionally, C6H6 and CO demonstrate higher contributions than temperature and humidity, consistent with prior studies indicating strong correlations between these pollutants and NOx [42], [43].

VI results from IMV-LSTM appear partially reasonable, with NOx and CO receiving high importance. Additionally, while the model captures the stepwise importance of earlier time points, its RNN-based architecture has limitations in modeling long-range dependencies. As the time lag increases, RNNs tend to assign uniform or diminished importance to distant time steps, complicating the accurate reflection of temporal influences from the past.

TFT provides reasonable contributions from NOx, NO2, and CO. However, the relatively lower contribution of NO2 compared to CO is inconsistent with domain knowledge. TI effectively captures influences from 10 to 13 h prior, recent time points, and the 24-hour periodicity, demonstrating the model's strong explainability.

MTEX yields expected results, with NOx dominating the VI and recent time points dominating TI. However, the model fails to capture the 24-hour periodic pattern and shows occasional spikes in TI that are difficult to interpret from a domain perspective.

Finally, XCM captures the impact of pollutant emissions 10 to 13 hours prior but neglects periodicity and shows a lower contribution of NO2 compared to other features.

These results indirectly emphasizes the superior forecasting performance of DLFormer on the PAQ dataset, as it effectively incorporates NO2, known for its strong correlation with NOx, into its predictions.

**b) PM2.5 Dataset Analysis**: In Beijing, PM2.5 concentrations and air quality indices are heavily influenced by meteorological conditions and human activities affecting atmospheric mixing. Thus, they typically exhibit a 24-hour periodicity, with additional immediate effects from wind, rain, and snow [44]-[46].

In Fig. 6, the VI results from the DLFormer show that snow (LS) and rain (LR) generally have lower absolute contributions due to their limited occurrences in the dataset. However, recent time points exert relatively greater influence when these events occur. Temperature (TEMP), dew point temperature (DEWP), and pressure (PRES) significantly impact atmospheric stability and circulation, with their influence primarily concentrated within the recent 12-hour window, consistent with existing literature. However, cumulative wind speed (LWS) tends to be underestimated. In Fig. 5, the TVA attention map for the PM2.5 dataset further supports the presence of this 24-hour cycle, with the model focusing on the periodic component and recent time intervals. PM2.5 values from 24 hours prior receive heavy weighting, while other variables also emphasize recent hours, reflecting a balance between long-term periodicity and short-term effects.

The VI results for IMV-LSTM indicate strong autoregressive contributions from PM2.5, similar to DLFormer, but with PRES ranking higher than TEMP, which contradicts domain knowledge. For TI, the model primarily captures short-term dependencies while largely neglecting the impact of LR.

TFT shows comparable VI patterns to DLFormer, with appropriate contributions from PM2.5, TEMP, and LR, while giving slightly higher importance to LR than DLFormer. In terms of TI, TFT effectively captures the 24-hour periodicity and recent contributions from LWS, LS, LR, and the target variable, demonstrating strong explainability.

For MTEX, VI results show uniform contributions across all variables, indicating limited differentiation. TI focuses on PM2.5 values from 22 hours prior, revealing partial periodicity. LS and LR show some recent influence, while other variables' temporal patterns are poorly distinguished.

Similarly, XCM captures the periodicity from approximately 22 hours prior and reflects a strong influence from recent LR values. However, it shows uniform temporal patterns across other variables, with LS, TEMP, and PRES exhibiting similar patterns contrary to domain expectations. Qualitative analysis demonstrates that DLFormer and TFT produce VI and TI patterns that are most consistent with domain knowledge across the two datasets. In contrast, IMV-LSTM exhibits partially consistent TI for short-term dependencies, but tends to converge toward the mean as the time distance from the prediction point increases, revealing its limitations in capturing long-term dependencies. Finally, the post-hoc explainable models, MTEX and XCM, reveal notable discrepancies compared to the existing literature. Both models fail to align with domain knowledge in terms of VI and TI, highlighting the inherent limitations of post-hoc explainability methods in accurately capturing complex multivariate time-series dynamics.

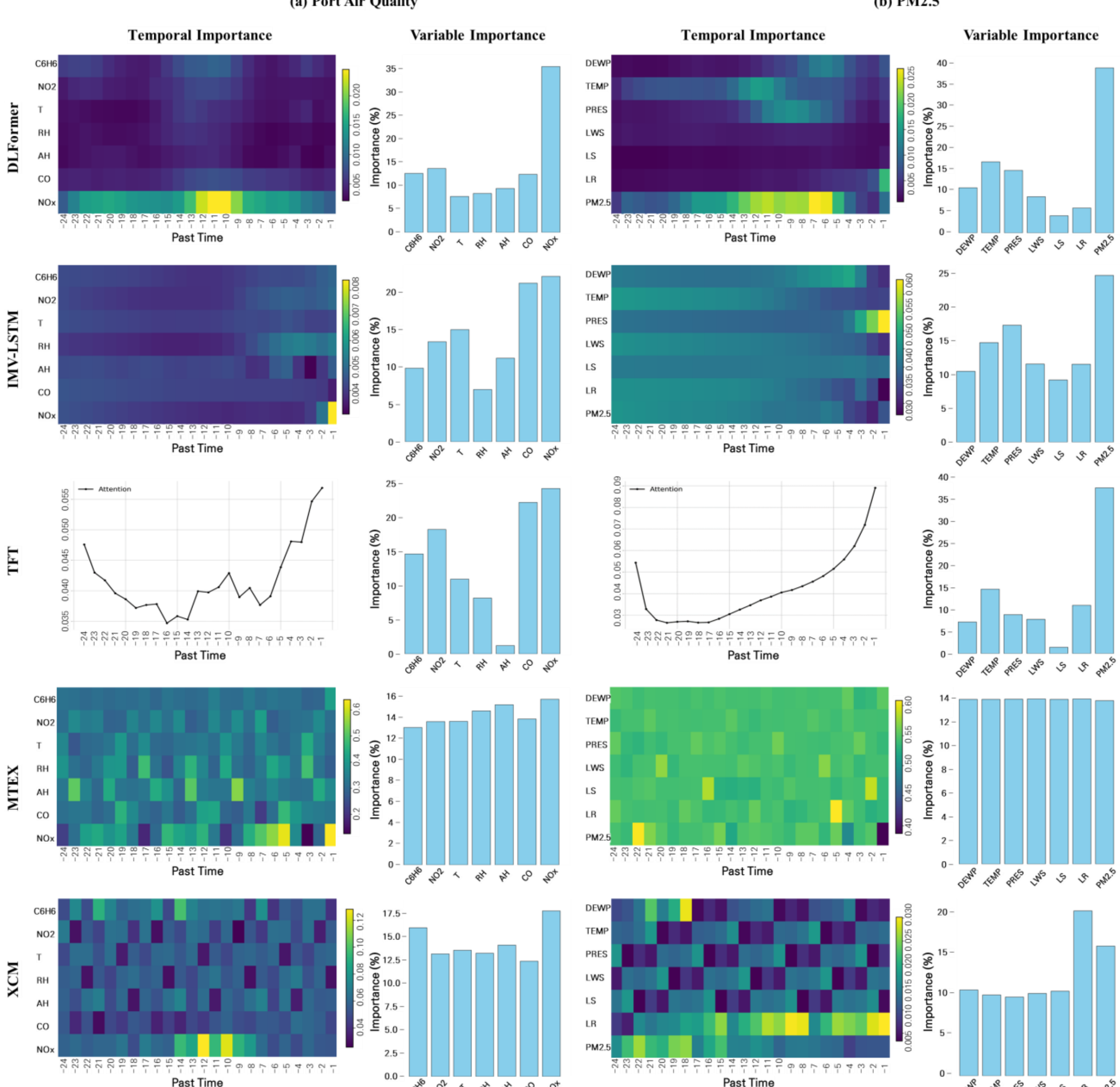

**Fig. 6.** Visualization comparing temporal importance and variable importance across different models for two datasets: (a) Port Air Quality and (b) PM2.5.

## V. Ablation Studies

We conducted an ablation study to evaluate the effectiveness of each component in DLFormer. Specifically, we address the following five research questions:

- Are the components of DL Embedding effective? (A)
- Is the inclusion of attention layers in DL Encoder and DL Decoder architectures justified? (B)
- Does scaling the parameter size of DLFormer to large language model (LLM) levels improve performance? (C)
- Does the TVA attention map effectively capture the temporal periodicity associated with time lags? (D)
- Can the TVA attention maps reasonably explain a single sample under specific conditions? (E)

### *A. Analysis of the Effects of DL Embedding Components*

To analyze the contribution of each component within DL Embedding, we conducted ablation experiments by selectively removing combinations of GSPE and LSPE. We first performed forecasting for all prediction lengths (1, 3, 6, 12, 96, 192, 336, and 720) and then calculated the average performance across all forecasting horizons. Table XIII presents the results of the component analysis of DL Embedding.

The values represent the average prediction results obtained from five independent experimental runs. The ablation study shows that DLFormer with GSPE and LSPE achieves the best performance, indicating that this embedding configuration serves as an optimized MTSF design.

TABLE XIII
DL EMBEDDING COMPONENTS EFFECT ANALYSIS RESULTS

| Case | GSPE+LSPE | | w/o GSPE | | w/o LSPE | |
|---|---|---|---|---|---|---|
| Metrics | MSE | MAE | MSE | MAE | MSE | MAE |
| PAQ | **0.876** | **0.642** | 0.966 | 0.682 | 0.925 | 0.663 |
| SML | **0.187** | **0.251** | 0.190 | 0.257 | 0.193 | 0.255 |
| Volatility | **0.383** | **0.374** | 0.484 | 0.411 | 0.443 | 0.393 |
| PM2.5 | **0.504** | **0.490** | 0.51 | 0.495 | 0.506 | 0.491 |
| Exchange | **0.096** | **0.196** | 0.111 | 0.203 | 0.098 | 0.197 |
| Weather | **0.160** | **0.242** | 0.19 | 0.277 | 0.162 | 0.256 |
| ETTh1 | **0.089** | **0.212** | 0.112 | 0.234 | 0.095 | 0.218 |
| ETTm1 | **0.067** | **0.165** | 0.068 | 0.166 | 0.069 | 0.168 |
| ETTh2 | **0.112** | **0.237** | 0.115 | 0.242 | 0.114 | 0.24 |
| ETTm2 | **0.068** | **0.161** | 0.09 | 0.183 | 0.078 | 0.172 |

TABLE XIV
ANALYSIS RESULTS OF ATTENTION OPERATION WITHIN DL ENCODER AND DL DECODER

| Enc / Dec | AL / AL | | LL / AL | | AL / LL | | LL / LL | |
|---|---|---|---|---|---|---|---|---|
| Metrics | MSE | MAE | MSE | MAE | MSE | MAE | MSE | MAE |
| PAQ | **0.876** | **0.642** | 0.877 | 0.647 | 0.896 | 0.648 | 0.948 | 0.671 |
| SML | **0.187** | **0.251** | 0.196 | 0.256 | 0.196 | 0.263 | 0.192 | 0.26 |
| Volatility | **0.383** | **0.374** | 0.478 | 0.404 | 0.552 | 0.445 | 0.387 | 0.382 |
| PM2.5 | **0.504** | **0.490** | 0.507 | 0.497 | 0.506 | 0.493 | 0.509 | 0.499 |
| Exchange | **0.096** | **0.196** | 0.098 | 0.197 | 0.116 | 0.209 | 0.098 | 0.198 |
| Weather | **0.160** | **0.242** | **0.160** | 0.248 | 0.218 | 0.290 | 0.165 | 0.255 |
| ETTh1 | 0.089 | 0.212 | 0.091 | 0.215 | **0.087** | 0.213 | 0.086 | **0.208** |
| ETTm1 | **0.067** | **0.165** | 0.071 | 0.171 | 0.068 | 0.166 | 0.069 | 0.169 |
| ETTh2 | **0.112** | **0.237** | 0.12 | 0.248 | 0.117 | 0.242 | 0.126 | 0.251 |
| ETTm2 | **0.068** | **0.161** | 0.108 | 0.195 | 0.074 | 0.167 | 0.082 | 0.176 |

### *B. Analysis of Attention Operation within DL Encoder and DL Decoder*

We experimented to compare the performance when replacing the attention layers in the original DL Encoder and DL Decoder with linear layers. Tables XIV presents the results of the attention operation. The values reflect forecasts for all prediction lengths (1, 3, 6, 12, 96, 192, 336, and 720) and the average performance across these horizons. These results confirm that the attention layers effectively capture the relationships in the input data when combined with DL Embeddings in the encoder and decoder networks. In contrast, the encoder-decoder structure with linear layers fails to effectively learn the interactions between temporal and variable dimensions. The weight-sharing nature of linear layers across multiple dimensions likely limits their ability to model complex cross-dimensional relationships when combined with DL Embeddings.

### *C. Analysis of DLFormer Scalability through LLM-Level Hidden Dimension Expansion*

Recent studies on MTSF have reported that large-scale models, such as TimeLLM [47], achieve strong predictive performance. Motivated by these findings, we further analyze the scalability of DLFormer by increasing its hidden dimension from 128 to 1024, matching the dimension used in TimeLLM based on GPT-2. Tables XV and XVI present the performance comparison results between the large-scale version of DLFormer (referred to as Large DLFormer) and TimeLLM. Scaling DLFormer to LLM-level hidden dimensions leads to an average performance improvement of 8.98%. This finding suggests that the scaling law also applies to DLFormer and demonstrates that the proposed embedding and learning strategies are highly effective, outperforming LLM-based models in MTSF tasks.

### *D. Analysis of the Time Lag Capturing Capability of the Time-Variable Aware Attention Map*

TVA attention maps serve as a key interpretability component in DLFormer, providing insights into the learned temporal dependencies for each prediction step. Specifically, it visualizes the attention weights across all past variables and time steps, contributing to each future prediction point. Similarly, the time-lagged cross-correlation (TLCC) is a statistical tool designed to quantify the temporal correlation between past variables and future prediction targets [48]. High TLCC values indicate that a specific time point of the given variable exhibits a strong correlation with the future horizon. Given this, if DLFormer effectively learns the intended variable-time relationships through DL Embedding, we hypothesize that the TVA attention map should exhibit patterns comparable to those observed in TLCC. Fig. 7 compares TVA attention maps and TLCC across multiple datasets. Subfigures (a)–(j) correspond to the PAQ, SML, Volatility, PM2.5, Exchange, Weather, ETTh1, ETTm1, ETTh2, and ETTm2 datasets, respectively. All comparisons were performed in the univariate prediction setting. For each dataset, the left panel displays the learned TVA attention map, while the right panel presents the corresponding TLCC heatmap for the same dataset. TLCC is visualized using the same axis configuration as the attention map, where the x-axis represents the time steps for each variable (with the rightmost position indicating the autoregressive target variable), and the y-axis denotes the forecast time step. The values reflect the average correlation across all samples for each (variable, time) pair. As expected in supervised forecasting tasks, the TVA attention map predominantly assigns high attention weights to past time steps of the target variable, indicating their strong predictive influence. Beyond this autoregressive effect, a notable observation emerges: for most datasets, except those with inherently stochastic characteristics such as exchange and volatility, the TVA attention map reveals patterns that closely align with the high-correlation regions in TLCC.

TABLE XV
PERFORMANCE COMPARISON BETWEEN LARGE DLFORMER AND TIMELLM WITH LLM-LEVEL HIDDEN DIMENSIONS

| | PAQ | | | SML | | | Volatility | | | PM2.5 | | | Exchange | | |
|---|---|---|---|---|---|---|---|---|---|---|---|---|---|---|---|
| Metrics | MSE | MAE | COR | MSE | MAE | COR | MSE | MAE | COR | MSE | MAE | COR | MSE | MAE | COR |
| DLFormer | 1.100 | 0.740 | 0.612 | 0.405 | 0.490 | 0.818 | 0.801 | 0.629 | **0.348** | 0.718 | 0.633 | 0.156 | 0.182 | 0.318 | 0.38 |
| Large DLFormer | **1.079** | **0.722** | **0.645** | **0.368** | **0.462** | **0.845** | **0.716** | **0.605** | 0.342 | **0.704** | **0.631** | **0.161** | **0.130** | **0.279** | **0.43** |
| TimeLLM | 1.257 | 0.826 | 0.532 | 0.417 | 0.517 | **0.845** | 0.749 | 0.636 | 0.105 | 0.908 | 0.688 | 0.131 | 0.294 | 0.381 | -0.02 |

TABLE XVI
PERFORMANCE COMPARISON BETWEEN LARGE DLFORMER AND TIMELLM WITH LLM-LEVEL HIDDEN DIMENSIONS

| | Weather | | | ETTh1 | | | ETTm1 | | | ETTh2 | | | ETTm2 | | |
|---|---|---|---|---|---|---|---|---|---|---|---|---|---|---|---|
| Metrics | MSE | MAE | COR | MSE | MAE | COR | MSE | MAE | COR | MSE | MAE | COR | MSE | MAE | COR |
| DLFormer | 0.293 | 0.396 | 0.430 | 0.149 | 0.300 | 0.198 | 0.124 | 0.263 | 0.218 | **0.182** | 0.337 | 0.584 | 0.121 | 0.261 | 0.657 |
| Large DLFormer | 0.277 | 0.383 | **0.432** | 0.134 | 0.288 | 0.213 | 0.099 | 0.238 | **0.240** | 0.183 | **0.335** | **0.608** | **0.117** | **0.252** | **0.676** |
| TimeLLM | **0.227** | **0.340** | 0.320 | **0.086** | **0.226** | **0.214** | **0.054** | **0.175** | 0.191 | 0.201 | 0.352 | 0.607 | 0.135 | 0.274 | 0.607 |

Moreover, similar to TLCC, the TVA attention map exhibits visually distinct grid-like structures that demarcate the temporal influence of each variable. This confirms that the DL Embedding framework effectively captures the variable-specific temporal relationships as intended. While prior studies [49], [50] have suggested that token embeddings in Transformers can degenerate into random noise, our results demonstrate otherwise. The comparison indicates that DLFormer, through its specialized DL Embedding and attention mechanisms, learns semantically meaningful representations that reflect variable-specific time dependencies critical for MTSF.

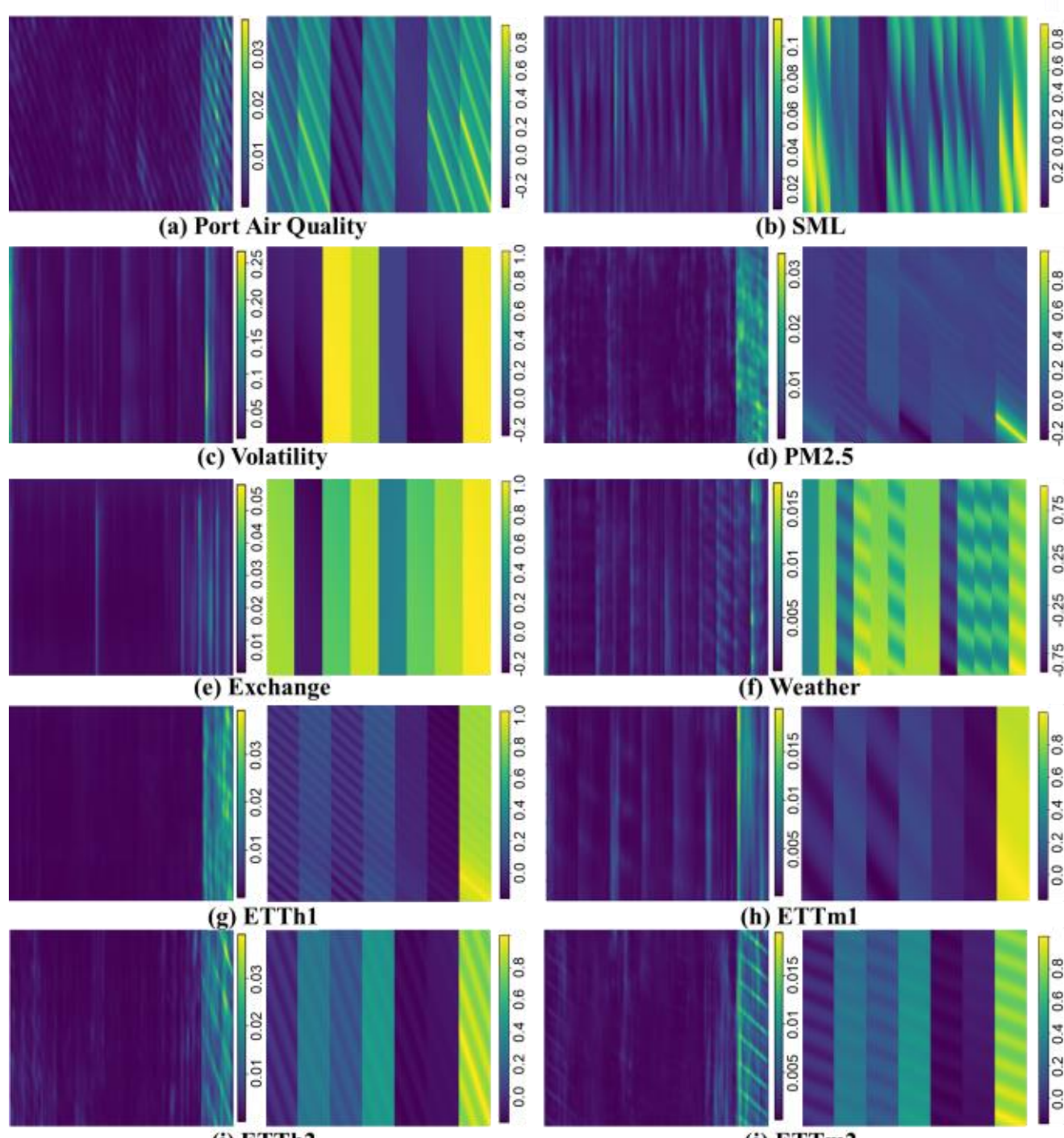

**Fig. 7.** Visualization of the comparison results between the time-variable-aware attention map and time lagged cross-correlation map.

### *E. Local Explainability Analysis of the Time-Variable Aware Attention Map*

To evaluate the local explainability of the TVA attention map, we conducted experiments using two datasets: (a) SML and (b) PM2.5, both characterized by distinct and observable changes under specific periods or conditions. The goal is to assess whether the TVA attention map can effectively highlight meaningful variations. Fig. 8 presents the local explainability analysis results for both datasets. The top row visualizes time-series plots for periods of significant change. The bottom-left quadrant illustrates the evolution of input signals across sequential samples and the corresponding TVA attention maps extracted by DLFormer. The bottom-right quadrant visualizes the predicted values of the target variable over the forecast horizon. All results are derived from samples drawn from the test set.

**a) SML Dataset**: In the SML dataset, the input length is set to 32 steps (8 h) and the prediction length to 16 steps (4 h). The input variable used for analysis is the Twilight Condition (TC). The top graph shows TC changing sharply between 5:00 a.m. and 6:00 a.m., aligning with sunrise. In Sample 1, the TVA attention map assigns high weights to the time step where TC begins to rise, precisely identifying the critical point. Subsequent samples consistently focus on this sunrise transition, correlating with a gradual increase in the predicted indoor temperature. This indicates that DLFormer successfully captures the temperature rise following sunrise, providing a meaningful and interpretable local explanation.

**b) PM2.5 Dataset**: In the PM2.5 dataset, the input and prediction lengths are set to 24 steps (24 h). The input variable is Hours of Rain (LR). The time-series plot indicates that rainfall begins around 16:00 on April 3rd, with the cumulative LR value increasing linearly thereafter. In Sample 1, TVA attention map accurately focuses on the moment rain begins. Consequently, the model predicts a decrease in PM2.5 concentration, consistent with established domain knowledge [46]. This pattern continues in subsequent samples, where the attention focuses on the onset and persistence of rainfall, leading to flattened, lower PM2.5 predictions that reflect the expected atmospheric cleansing effect. These findings suggest that DLFormer captures the impact of specific input variables and provides locally interpretable explanations for its forecasts.

These results confirm that the TVA attention map demonstrates strong local explainability, accurately identifying variable-specific temporal effects under specific conditions and producing interpretable attention behaviors aligned with real-world dynamics.

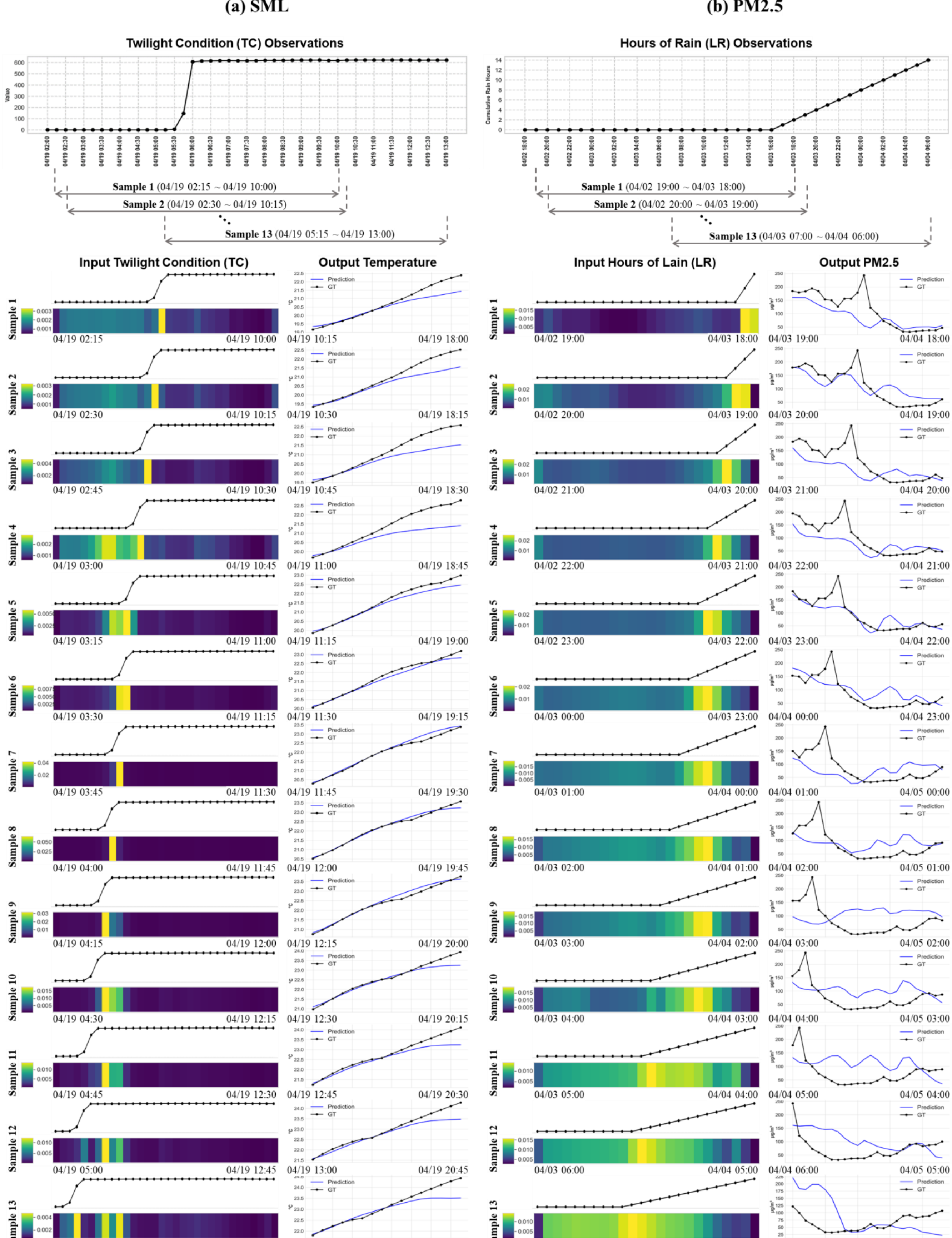

**Fig. 8.** Detailed visualization of the time-variable-aware attention map for local explainability in (a) SML, (b) PM2.5 data.

## VI. CONCLUSION

In this study, we presented DLFormer, a novel Transformer-based architecture tailored for explainable multivariate time series forecasting (MTSF) in big data environments. DLFormer introduces the Time-Variable-Aware Learning (TVAL) mechanism by integrating a Distributed Lag (DL) Embedding that structurally captures both temporal and variable-wise dependencies inherent in large-scale, high-dimensional time series data. By explicitly modeling lagged effects across variables, DLFormer achieves scalable forecasting performance while significantly enhancing model explainability from both global and local perspectives—addressing critical needs in real-world big data analytics.

Comprehensive experiments on benchmark and real-world datasets demonstrated DLFormer's superior performance compared to existing explainable models, achieving consistent improvements in MSE, MAE, and correlation metrics across various forecasting horizons. Furthermore, the qualitative and quantitative evaluations of the proposed TVA attention map confirm that DLFormer provides robust and interpretable insights into how historical variable patterns influence predictions.

This study bridges the gap between forecasting performance and explainability in DL-based time series forecasting. Future work includes extending the DLFormer framework to handle irregular time series data and incorporating domain-adaptive mechanisms to further enhance generalizability and robustness in diverse real-world applications.

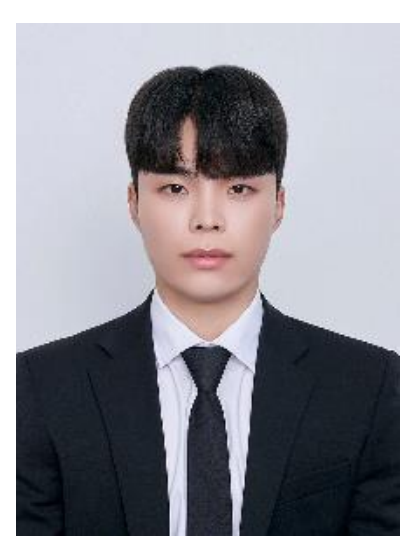

**Youngwhi Kim** received his B.S. in Creative Software Engineering from Dong-eui University, Republic of Korea, in 2024, where he is currently pursuing M.S. in Graduate School of AI Convergence Engineering from the Changwon National University, Republic of Korea. His research interests include data mining and deep learning, with a particular focus on time-series neural network structure design and explainable artificial intelligence.

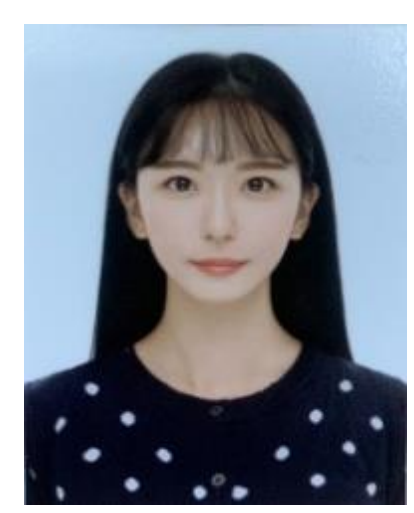

**Dohee Kim** (Member, IEEE) received her B.S. in Industrial Engineering from Pusan National University, Republic of Korea, in 2019 and her M.S. and Ph.D. in Industrial Engineering from Pusan National University, Republic of Korea, in 2024. She is currently a research professor at the Safe & Clean Supply Chain Research Center, Pusan National University, Republic of Korea. Her research interests include deep learning, optimization, representation learning, and datamining.

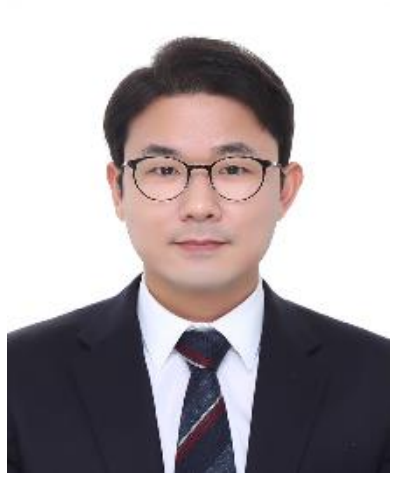

**Joongrok Kim** received his M.S. in Graduate Program in Biometrics and his Ph.D. degree in Electrical and Electronic Engineering from Yonsei University, Republic of Korea, in 2008 and 2014, respectively. From 2014 to 2024, he was a researcher at the LG Electronics AI Lab, where he worked on applying artificial intelligence technologies to various domains including automotive and home appliances. He is currently an associate professor with the Department of Artificial Intelligence Convergence Engineering at Changwon National University, Republic of Korea. His research interests include computer vision, deep learning, time series forecasting, and AI applications in industrial systems.

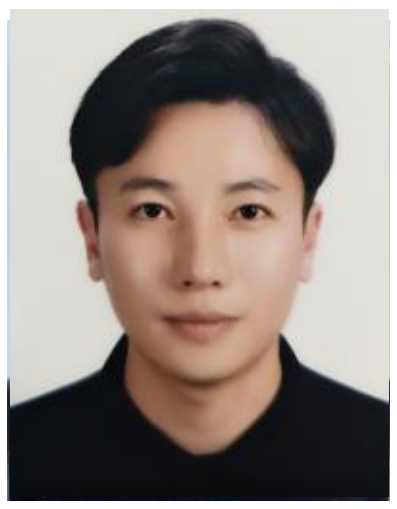

**Sunghyun Sim** (Member, IEEE) received his B.S. in Statistics from Pusan National University, Republic of Korea, in 2016 and his Ph.D. in Industrial Engineering from Pusan National University, Republic of Korea, in 2021. He is currently an assistant professor with the Department of Artificial Intelligence Convergence Engineering at Changwon National University, Republic of Korea. His research interests include data mining, process mining, and deep learning, with a particular focus on neural layer structure design, lightweight deep learning models, and explainable artificial intelligence.